\documentclass[10pt,journal,letterpaper]{IEEEtran}

\usepackage[utf8]{inputenc}
\usepackage[T1]{fontenc}
\usepackage{amsmath,amssymb,amsfonts}
\usepackage{graphicx}
\usepackage[hyphens]{url}
\usepackage{cite}
\usepackage{booktabs}
\usepackage{nicefrac}
\usepackage{microtype}
\usepackage{xcolor}
\usepackage{csquotes}
\usepackage[load-configurations=version-1]{siunitx}
\usepackage{tikz}
\usetikzlibrary{arrows.meta,positioning}
\usepackage{adjustbox}
\usepackage{enumitem}
\usepackage{longtable}
\usepackage{etoolbox}
\usepackage[hidelinks]{hyperref}

\AtBeginEnvironment{figure}{\let\textwidth\columnwidth}
\makeatletter
\@ifundefined{endonecolumn}{}{}
\makeatother

\newif\ifFullSizeMainFigures
\newif\ifPublicArtifacts

\FullSizeMainFigurestrue
\PublicArtifactstrue

\newcommand{\NSAISurveyIndex}{\href{https://brandonio-c.github.io/NSAI-2025-Survey/}{NSAI survey and audit index}}
\newcommand{\InteractiveResultsBrowser}{\href{https://brandonio-c.github.io/NSAI-2025-Survey/}{interactive results browser}}

\newcommand{\PublicExtractionSheet}{\href{https://docs.google.com/spreadsheets/d/1ueACvF21PErkg4qKM3xdhPFmnwNvHR6r5eDWG_lMXDI/edit}{public reproduction and data-extraction sheet}}
\newcommand{\ReproductionLog}{\href{https://docs.google.com/spreadsheets/d/1ueACvF21PErkg4qKM3xdhPFmnwNvHR6r5eDWG_lMXDI/edit?usp=sharing}{NSAI reproduction and data-extraction results}}

\newcommand{\AuditCluster}{the University of Maryland Zaratan high-performance computing cluster}
\newcommand{\AuditFileSystem}{the University of Maryland Zaratan project scratch file system}
\newcommand{\AuditArchive}{\url{https://doi.org/10.5281/zenodo.21779234}}
\newcommand{\AnnotatorTrainingMaterials}{annotator training and calibration materials in the \AuditArchive}
\newcommand{\DataExtractionDemo}{\href{https://www.youtube.com/watch?v=8SY8VoDUliU}{NSAI-2025 data-extraction worked example}}

\title{6.5\% of the Neuro-Symbolic Literature Can Be Reproduced from Its Published Artifacts, a Six-Stage Audit Framework and First Instantiation}

\author{Brandon Colelough, Vladimir Martirosyan, Ishan Tamrakar, William Regli, Aditya Kumar, Anh N. Nhu, Dhruv Dubey, Raj Ambavane, and Haowei Deng%
\thanks{The authors are with the Department of Computer Science, University of Maryland, College Park, Maryland, USA. Corresponding author: Brandon Colelough (brandcol@umd.edu).}}

\begin{document}
\maketitle

\begin{abstract}
We present a six-stage framework for auditing the reproducibility of scientific claims across a research literature within the computer science domain, and instantiate our framework for the neuro-symbolic AI (NSAI) subdomain. We applied the framework during a nine-month audit of studies spanning multiple publication years. Stage one retrieved \num{5497} records and removed \num{3018} duplicates. Stage two screened the \num{2479} unique records at title and abstract, identifying \num{1365} self-identified NSAI records, then removed a further \num{61} at full text for off-topic, non-research, no-quantitative-evaluation, or inaccessible-full-text reasons. Stage three sought a verifiable public code artifact for each of the \num{1304} eligible records and found none for \num{849}, leaving \num{455} attempted reruns. Stage four assessed executable environments, stage five attempted the primary experiments, and stage six extracted and audited the outcomes. We fully reproduced \num{48} studies and partially reproduced \num{37}; their combined count of \num{85} represents \SI{6.52}{\percent} of the eligible corpus and \SI{18.68}{\percent} of attempted reruns. We classified \num{321} attempts under missing, incomplete, or environment-related artifact failure categories, \num{42} under missing or unusable code repositories, and \num{7} under failure despite nominally complete artifacts. These figures quantify a persistent reproducibility deficit that survives even nominal \enquote{code available} declarations, and signal the need for enforced, versioned, and permanently archived artifact bundles in future NSAI publications. We argue that empirical NSAI papers should be required at submission time to provide complete, versioned, and permanently archived artifact bundles. 
\end{abstract}


\section{Introduction}
Can an independent team rerun the computational pipeline released with a published paper and recover its main reported result? We put that question to 455 neuro-symbolic studies that advertised a public code artifact. Of these, 48 met the full-reproduction criterion and 37 met the partial-reproduction criterion.  In this paper, we undertake a study of the same-artifact rerun of published results in the emerging field of Neuro-Symbolic Artificial Intelligence. Specifically, we ask whether an independent team can rerun the computational pipeline released with a paper and recover the main reported result within a prespecified tolerance. NSAI is viewed by many as the emerging frontier that merges concepts from connectionist methods with those referred to informally as \enquote{Good Old Fashioned AI}, consisting of logic and formal techniques for reasoning. Integrations range from loose pipelines, where a neural component feeds a symbolic one, to tightly coupled systems trained jointly. The frontier science that aims to integrate these two techniques is highly dynamic, and while the number of papers that claim to be creating contributions in this area is exploding in number, this paper reports that the results in the majority of these papers cannot be recovered by rerunning the released artifacts. This indicates that the research area comprising \enquote{Neuro-Symbolic AI} is experiencing a deficit in same-artifact rerun reproducibility and lacks the community norms that would make released artifacts reliably executable as part of the publication of work.  For the field to advance into the position of critical importance that many feel is inevitable, we should raise the bar on our scientific standards and require greater emphasis on how we can document reproducibility so that others may verify our claims. On the evidence below, we argue that empirical submissions in this field should carry a complete, versioned, and permanently archived artifact bundle at submission time.

\section{Background}
As has been documented often in the recent literature, many areas of science are currently experiencing a \enquote{reproducibility crisis} \cite{Baker2016-vl} as studies have proven difficult or impossible to replicate \cite{Lopez-Nicolas2022-pp}, data sources are not available \cite{Milkowski2018-kh}, and experimental assumptions are not made explicit \cite{Hensel2020-nm}. Social sciences, in which these phenomena have been documented most extensively, have the additional issues posed by post hoc redesign of scientific hypotheses. Known as \enquote{Hypothesizing After the Results are Known}, or \enquote{p-hacking}, this occurs when a researcher forms or rewrites a hypothesis after seeing the data, and then presents that hypothesis as if it were specified before the data were collected \cite{Rubin2022-wa}. The field of computing, in theory, should be highly reproducible, as algorithms, code, data, and other artifacts can be easily shared and adopted.  Conferences and journals have begun to require data sharing and other best practices to improve reproducibility \cite{noauthor_2020-cv}.  Without reproducible artifacts, computational research risks becoming unfalsifiable claims that cannot be independently verified and thus fall outside the bounds of science. We need to be able to rigorously and independently test hypotheses laid out in papers in the computer science community in the same way we do in other domains, so that the work can have more credibility.


\subsection{Contributions and Research Questions}
We follow the ACM badging terminology \cite{acm2020badging} for reproducibility and refer to reproducibility as meaning that a team other than the original obtains a consistent result using the original team's artifacts. We make two contributions. The first is a six-stage framework for auditing same-artifact rerun reproducibility across a literature. The second is an instantiation of the audit framework itself, the largest of its kind, employing the framework on \num{1304} eligible NSAI records. 
The research questions we aim to address include: \textbf{\emph{RQ1}} What proportion of NSAI papers releasing code can be fully or partially reproduced? \textbf{\emph{RQ2}} How does the availability of code, data, and model weights, within the broader six-component artifact inventory, relate to rerun outcomes? 
\textbf{\emph{RQ3}} Do reproduction outcomes vary systematically by publication year or venue family (conference, journal, preprint)?

\subsection{Related Work}
\label{subsec:trends-patterns}

Reproducibility failures are well-documented across empirical disciplines. Landmark studies in psychology and medicine have shown that a substantial fraction of published findings cannot be independently replicated \cite{Open_Science_Collaboration2015-zt, Ioannidis2005-tf}, and large-scale surveys across the social and biomedical sciences trace these failures to non-disclosed analytical choices, unavailable data, and selective reporting \cite{Baker2016-vl, Lopez-Nicolas2022-pp, Hensel2020-nm}. The interested reader is referred to that broader literature for a full treatment. Computer science presents a structurally different case, as algorithms, code, and data can in principle be shared exactly, meaning empirical claims ought to be among the most verifiable in science. In practice, however, the same failure modes recur, as the studies below illustrate. Vanderdonckt and Vatavu introduced \emph{Amplitum}, a contextual framework that augments generic replication taxonomies with explicit descriptors of participants, devices, and physical settings \cite{10.1145/3736731.3746142}. Their gesture-elicitation case study reproduced prior findings only after replicators matched the original laboratory environment and user cohort, underscoring that code and stimuli alone are insufficient when human behavior is the main dependent variable. Ferrari Dacrema \emph{et al.} inspected 26 \enquote{state-of-the-art} (for the time) neural recommender papers (2015–2018) and were able to fully reproduce only 12 of the original 26, and in doing so discovered that \emph{eleven of those twelve} fell behind well-tuned neighborhood, matrix-factorization, or sparse-linear baselines once a common experimental protocol was enforced \cite{10.1145/3434185}. Ferrari et al. attribute the apparent performance gains to methodological weaknesses, including baselines left at default settings, inadvertent test-set leakage during epoch selection, and undocumented data-split choices.  The ReproNLP shared task series asks independent teams to repeat published human evaluations using materials supplied by the original authors, and reports that agreement with the original figures is frequently not achieved \cite{belz-thomson-2023-2023}. The organisers attribute divergence to underspecified evaluation protocols and to details of the original setup that were never documented. Henderson \textit{et al.} evaluated several widely cited DRL algorithms and found that benchmark rankings could reverse when random seeds, hardware platforms, or training horizons were varied \cite{10.5555/3504035.3504427}. Their study therefore recommends reporting results over many seeds, applying formal significance tests, and disclosing every experimental detail. Pawlik \emph{et~al.} examined the longevity of public datasets and found that many links degrade, move, or silently change versions, rendering later experiments irreproducible despite nominal accessibility \cite{10.1007/s13222-019-00317-8}. They argue for immutable storage, rigorous version identifiers, and provenance metadata. Reproduction breaks down whenever any piece of the experimental context is missing, and Neuro-Symbolic AI is no exception. An open-code link without hardware details, solver commits, pre-processing scripts, dataset splits, and hyperparameter schedules offers little more than performative compliance. Gains attributed to the symbolic–neural fusion may vanish once baselines receive equal tuning or a knowledge base is revised; stochastic variation in the neural component could dominate the symbolic layer as well. Reliable reproduction, therefore, requires immutable, versioned artifact bundles, multi-seed evaluation, and full provenance for pretrained weights, logic programs, and curated knowledge graphs.

\section{Methodological Audit Framework}

\ifFullSizeMainFigures
  \def\prismascale{0.9}
\else
  \def\prismascale{0.7}
\fi

\begin{figure}[t]
\centering
\begin{tikzpicture}[
scale=\prismascale,
transform shape,
  node distance = 8mm,
  box/.style   = {rectangle, draw, rounded corners,
                  minimum width=42mm, minimum height=10mm,
                  align=center, font=\scriptsize},
  side/.style  = {box, minimum width=46mm},
  outcome/.style = {rectangle, draw, rounded corners, dashed,
                    minimum width=42mm, minimum height=10mm,
                    align=center, font=\scriptsize},
  arr/.style   = {-{Stealth[length=2.6mm,width=2.3mm]}}
]

\node[box] (id)   {Records identified\\from databases\\\textbf{n = \num{5497}}};
\node[box, below=of id] (uniq){Unique records after duplicates\\\textbf{n = \num{2479}}};
\node[box, below=of uniq](scr) {Records screened\\(title/abstract)\\\textbf{n = \num{2479}}};
\node[box, below=of scr] (topic){Records on topic\\\textbf{n = \num{1365}}};
\node[box, below=of topic](elig){Eligible full-text NSAI records\\\textbf{n = \num{1304}}};
\node[box, below=of elig](attempt){Attempted reruns\\\textbf{n = \num{455}}};
\node[outcome, below=of attempt](repr){Reproduced and included\\\textbf{n = \num{85}}};
\node[side, right=10mm of attempt] (nrep){Not successfully\\reproduced\\\textbf{n = \num{370}}};

\draw[arr] (attempt.east) -- ++(8mm,0) |- (nrep.west);

\foreach \a/\b in {id/uniq,uniq/scr,scr/topic,topic/elig,elig/attempt}
  {\draw[arr] (\a.south) -- (\b.north);}

\draw[arr, dashed] (attempt.south) -- (repr.north);

\node[side, right=10mm of id]   (dup){Duplicates removed\\\textbf{n = \num{3018}}};
\node[side, right=10mm of scr]  (ex1){Title/abstract\\exclusions\\\textbf{n = \num{1114}}};
\node[side, right=10mm of topic](ex2){Full-text eligibility\\exclusions\\\textbf{n = \num{61}}};
\node[side, right=10mm of elig] (ex3){No verifiable public\\code artifact\\\textbf{n = \num{849}}};

\draw[arr] (id.east)   -- ++(8mm,0) |- (dup.west);
\draw[arr] (scr.east)  -- ++(8mm,0) |- (ex1.west);
\draw[arr] (topic.east)-- ++(8mm,0) |- (ex2.west);
\draw[arr] (elig.east) -- ++(8mm,0) |- (ex3.west);

\end{tikzpicture}
\caption{Selection and eligibility flow for the NSAI audit. Late full-text eligibility exclusions are shown separately and are not counted as rerun outcomes. The dashed bottom box reports the number of successful reruns within the attempted set and is shown as an audit outcome.}
\label{fig:prisma}
\end{figure}

We present our six-stage framework for auditing same-artifact rerun reproducibility across a literature and a realised example of our six-stage reproducibility framework below. Stage 1 constructs and screens the corpus. Stage 2 confirms full-text eligibility. Stage 3 identifies the code artifact and records a six-item artifact inventory before any execution is attempted. Stage 4 builds the released environment and checks executable integrity. Stage 5 reruns the primary experiment under a bounded repair allowance and a stated fidelity criterion. Stage 6 extracts data and audits outcome labels. 

\subsection{Corpus Construction and Screening (Stages 1--3)}
\label{subsec:corpus-screening}
Stage 1 constructs the corpus. We began with a deliberately broad bibliographic sweep, designed to capture the full breadth of the neuro-symbolic domain. Guided by a PRESS-validated query \cite{McGowan2016-zk} centered on \enquote{neuro-symbolic OR NeSy OR NSAI}, we queried nine major digital libraries on 23 May 2025: \textsc{Web~of~Science}, \textsc{Scopus}, \textsc{PubMed}, \textsc{Ei~Compendex}, \textsc{IEEE~Xplore}, \textsc{ACM~DL}, \textsc{SpringerLink}, \textsc{Google~Scholar}, and \textsc{arXiv}. The query returned \num{5\,497} records. This broad sweep increased coverage of emerging or lexically idiosyncratic work that narrower queries often overlook. A de-duplication pipeline removed \num{3\,018} duplicates (55\% of raw hits), leaving \num{2\,479} unique records. The substantial overlap across sources illustrates the cross-posting norm in NSAI research and underscores the need for multi-pass de-duplication.\footnote{Pipeline tools: EndNote, Covidence, Zotero, SR-Accelerator, and Rayyan.} Relevance screening was performed in a single-blind title-and-abstract pass. After a 20-record calibration, reviewers screened titles and abstracts and retained any paper whose authors explicitly described the work as \enquote{neuro-symbolic,} regardless of application domain. \num{1\,365} papers satisfied this broad criterion. Code availability was not inferred from the abstract alone. Instead, repository verification was conducted after full-text eligibility assessment, to determine entry into the attempted-rerun set, wherein annotators first inspected the full paper for repository or artifact links, and when none were present, they performed a structured external search using the paper title, author names, and method or domain keywords. This separate artifact-availability gate identified \num{849} eligible records without a verifiable public code artifact and left \num{455} studies for the rerun audit, an attrition of \SI{65.11}{\percent} that exposes the gap between open-science claims and delivered artifacts. The stages reduced the literature from \num{2\,479} unique records to \num{1\,365} NSAI-relevant papers and further to \num{455} audit candidates that at minimum linked or could be matched to a code artifact. Figure~\ref{fig:prisma} visualizes each reduction step, linking corpus construction directly to our objective of quantifying verifiable progress in Neuro-Symbolic AI.

\subsection{Eligibility Rules and Audit Outcomes}
\label{subsec:replication-protocol}

\subsubsection{Eligibility and attempted-rerun entry criteria}
For full-text eligibility, records were required to describe original empirical
neuro-symbolic research, provide quantitative evaluation relevant to computer
science, and have accessible full text. These requirements correspond to
$IC_1$ (self-described neuro-symbolic integration), $IC_2$ (empirical
evaluation), $IC_3$ (computer-science relevance), and $IC_5$ (full-text
availability). The additional $IC_4$ requirement---a directly linked or
uniquely identifiable public code artifact---was applied \emph{after}
full-text eligibility, to select the attempted-rerun cohort. Thus,
\num{1304} studies were eligible at full text, but only \num{455}
passed the code-artifact gate; the other \num{849} were retained
in the eligible-corpus denominator but were not counted as attempted reruns.

\subsubsection{Exclusion and failure accounting}
The full-text eligibility exclusions were: $E_1$, non-English language;
$E_2$, review/editorial/non-original empirical research;
$E_5$, no quantitative evaluation; $E_6$, outside the neuro-symbolic scope;
$E_7$, duplicate or superseded record; and $E_8$, inaccessible full text.
The source annotation vocabulary also included $E_3$ (no verifiable public
code) and $E_4$ (missing indispensable artifacts). For Figure~\ref{fig:prisma},
$E_3$ refers to the separate post-eligibility code-artifact gate rather
than to one of the 61 late full-text exclusions. After a record entered
the attempted-rerun cohort, any inaccessible indispensable inputs were
classified as an $O_4$ outcome, not as a full-text exclusion; environment
or code execution failures were separately classified as $O_5$.

\subsubsection{Reproduction Procedure}
Post-entry rerun outcomes are reported in Figure~\ref{fig:funnel-overall-pie}. Every study with a publicly accessible repository was evaluated under the six-stage protocol that compressed the workflow into discrete, auditable checkpoints \footnote{\NSAISurveyIndex}. All annotators began with a one-hour onboarding workshop\footnote{\AnnotatorTrainingMaterials} that introduced the audit workflow and extraction template. This was followed by two 4-hour live sessions (covering environment builds and dependency management, metric verification, and licensing constraints) and a 3-hour supervised drop-in lab in which each participant could reproduce exemplar studies end-to-end in a supervised environment. A worked example demonstrating application of the study's
data-extraction form is also available.\footnote{\DataExtractionDemo} All rerun attempts and artifact assessments were conducted by a team of eight trained graduate student annotators over a nine-month period, who completed the study’s standardized onboarding workshop and calibration protocol. Each annotator handled an average of 64.5 records, comprising the \num{455} attempted reruns and the 61 late eligibility exclusions distributed across eight annotators. Workload was allocated randomly across the team. A separate ten-paper rerun-annotation calibration pilot (distinct from the 20-record title-and-abstract screening calibration) produced Cohen’s $\kappa = 0.82$ (a standard 
measure of inter-rater agreement, where 1.0 is perfect agreement and values 
above 0.80 are considered strong), and outstanding disagreements were reconciled in group discussion. During the nine-month audit phase, the team met every second week to review edge cases and realign on the reproduction protocol. Outcome labels were assigned after entry into the attempted rerun pool as follows: \textbf{O1) }\textbf{Fully reproduced.} The primary experiment was executed successfully, and the reproduced result satisfied the study’s fidelity criterion. \textbf{O2) }\textbf{Partially reproduced.} The core pipeline executed, and the paper’s main qualitative claim was preserved, but one or more quantitative results fell outside the acceptance band, or only a subset of the headline experiments could be rerun. \textbf{O3) }\textbf{Executed but did not reproduce within tolerance.} The system ran to completion, but the reproduced results materially exceeded the acceptance threshold or contradicted the paper’s main quantitative claim. \textbf{O4) }\textbf{Not executable due to missing or inaccessible artifacts.} The attempted rerun could not proceed because one or more indispensable artifacts were unavailable, inaccessible, or not reconstructable under the study protocol. \textbf{O5) }\textbf{Not executable due to environment or code failure.} Attempted rerun failed because the released code or environment could not be built or executed under the protocol despite the permitted minimal fixes. For this study, an artifact was classified as missing only when it was an indispensable input to the primary experiment and could not be reconstructed from the paper and released materials under the study protocol. Indispensable inputs included fixed datasets or splits, preprocessing scripts that alter data semantics, checkpoints or weights when evaluation depended on a fixed trained model state, rule sets or knowledge bases, configuration files, environment descriptors, and evaluation assets.  All reproduction logs and extracted data are version-controlled, and a consolidated record is available in the \PublicExtractionSheet.

\subsection{Evaluation Design}
\label{subsec:evaluation-design}

We distinguish two successful outcomes. Full reproduction ($O_1$) is the strict criterion, requiring recovery of the paper's headline result; partial reproduction ($O_2$) is reported separately. The headline full-or-partial rate combines $O_1$ and $O_2$, while their individual rates remain distinguishable. Only papers that entered the attempted rerun pool count toward either. To qualify as a full reproduction, a repository must execute under the author-supplied environment (or a minimally updated equivalent), with a max 7 day wall-clock time and on the audit team's GPU pool, and reproduce the paper’s primary metric to within $\pm5$\,\% absolute error (or inside the authors’ 95\,\% confidence interval), yield outputs consistent with the paper’s headline claims, and require no correction of bugs intrinsic to the model architecture. Annotators were authorized to (i) update deprecated package versions or apply path fixes, provided it is minimal, (ii) adjust file paths, (iii) supply a lightweight evaluation harness when none was provided, and (iv) patch minor scripting errors.  Malformed or undocumented environments, missing indispensable post-entry artifacts, or architecture-level defects were recorded as O4 or O5 audit outcomes, not as exclusions.

\ifFullSizeMainFigures
\begin{figure*}[t]
\centering
  \includegraphics[width=\textwidth]{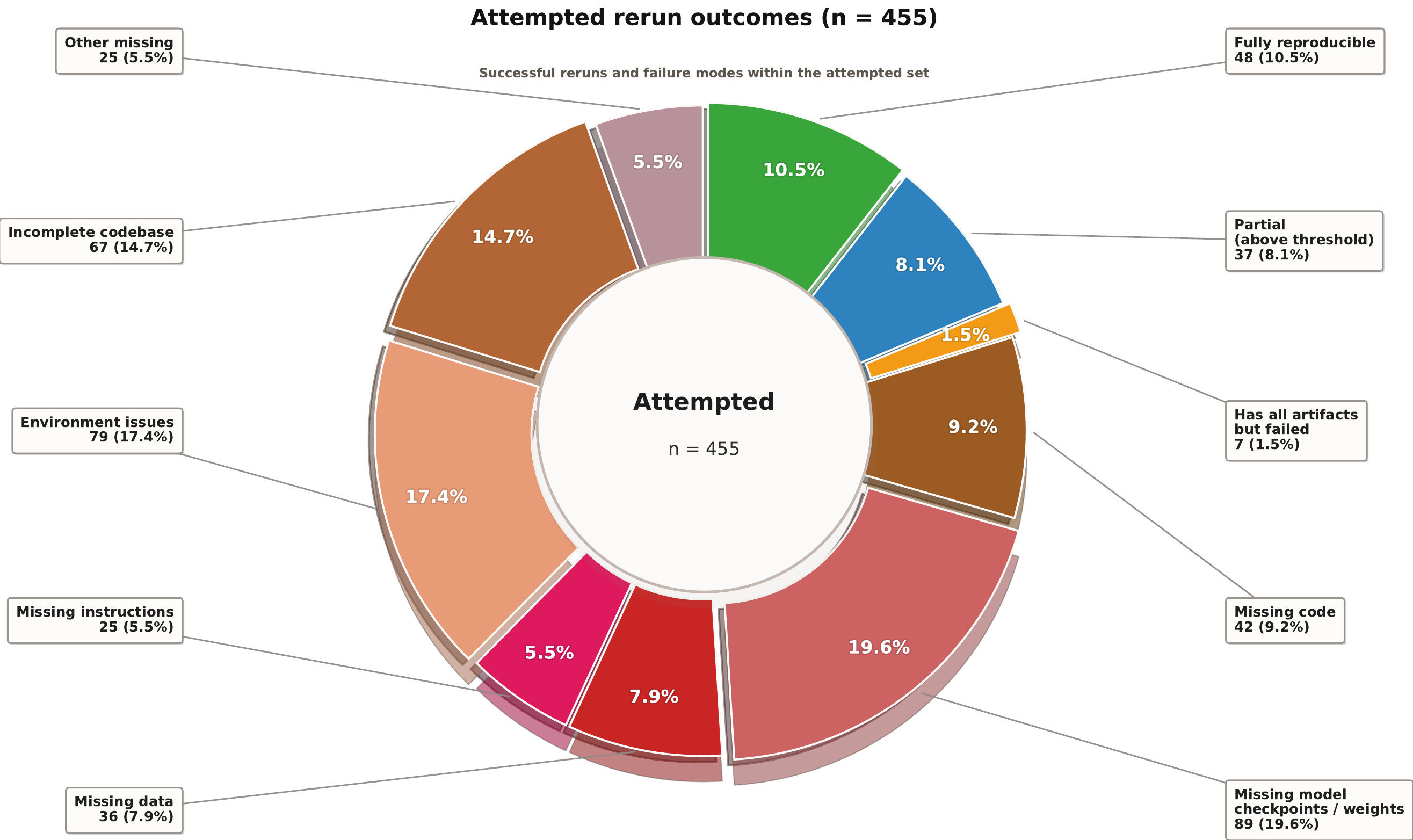}
  \caption{Rerun outcomes for the attempted same-artifact audit set (\(n = 455\)).}
  \label{fig:funnel-overall-pie}
\end{figure*}
\else
\begin{figure}[t]
\centering
  \includegraphics[width=\textwidth]{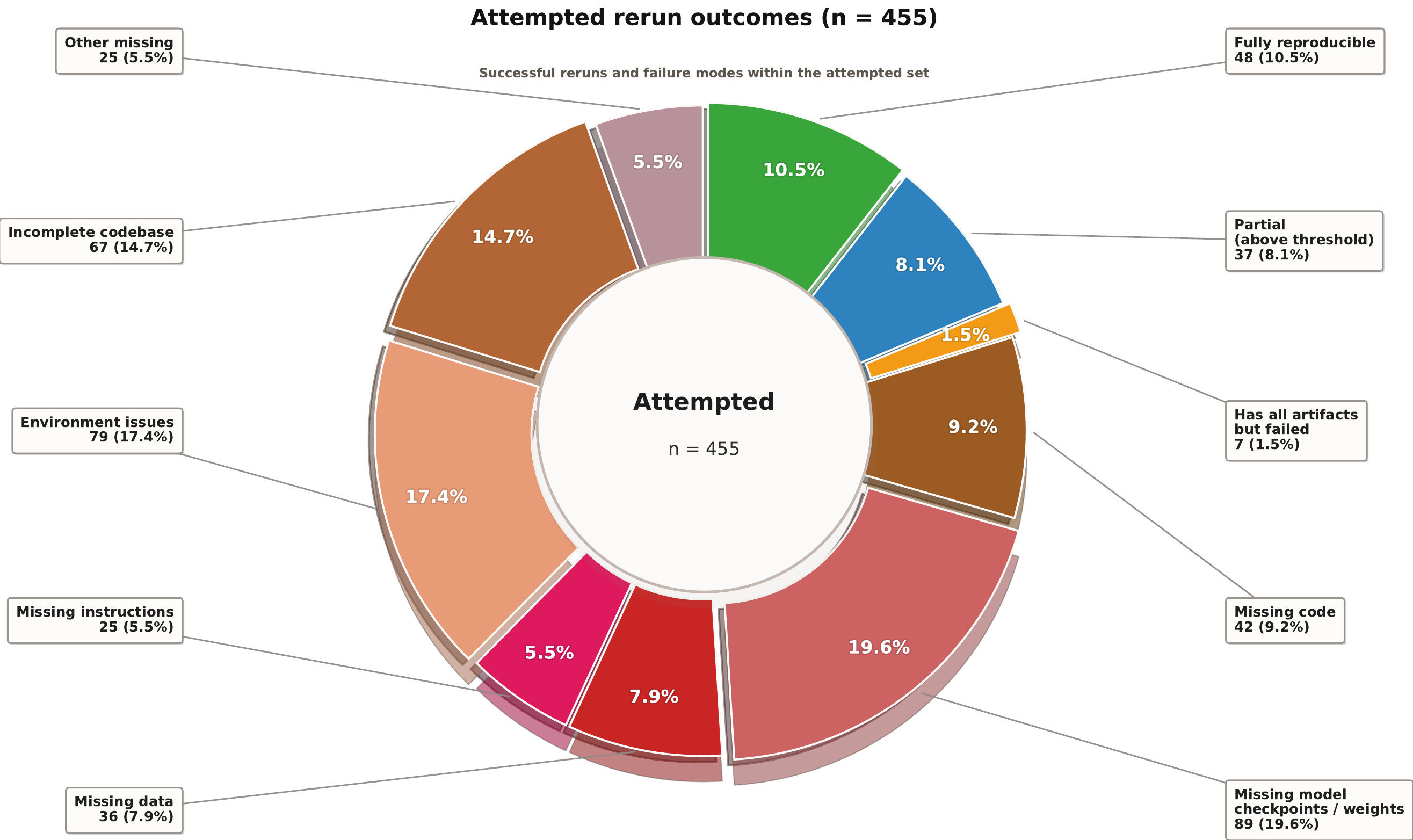}
  \caption{Rerun outcomes for the attempted same-artifact audit set (\(n = 455\)).}
  \label{fig:funnel-overall-pie}
\end{figure}
\fi

\ifFullSizeMainFigures
\begin{figure*}[t]
\centering
    \includegraphics[width=\textwidth]{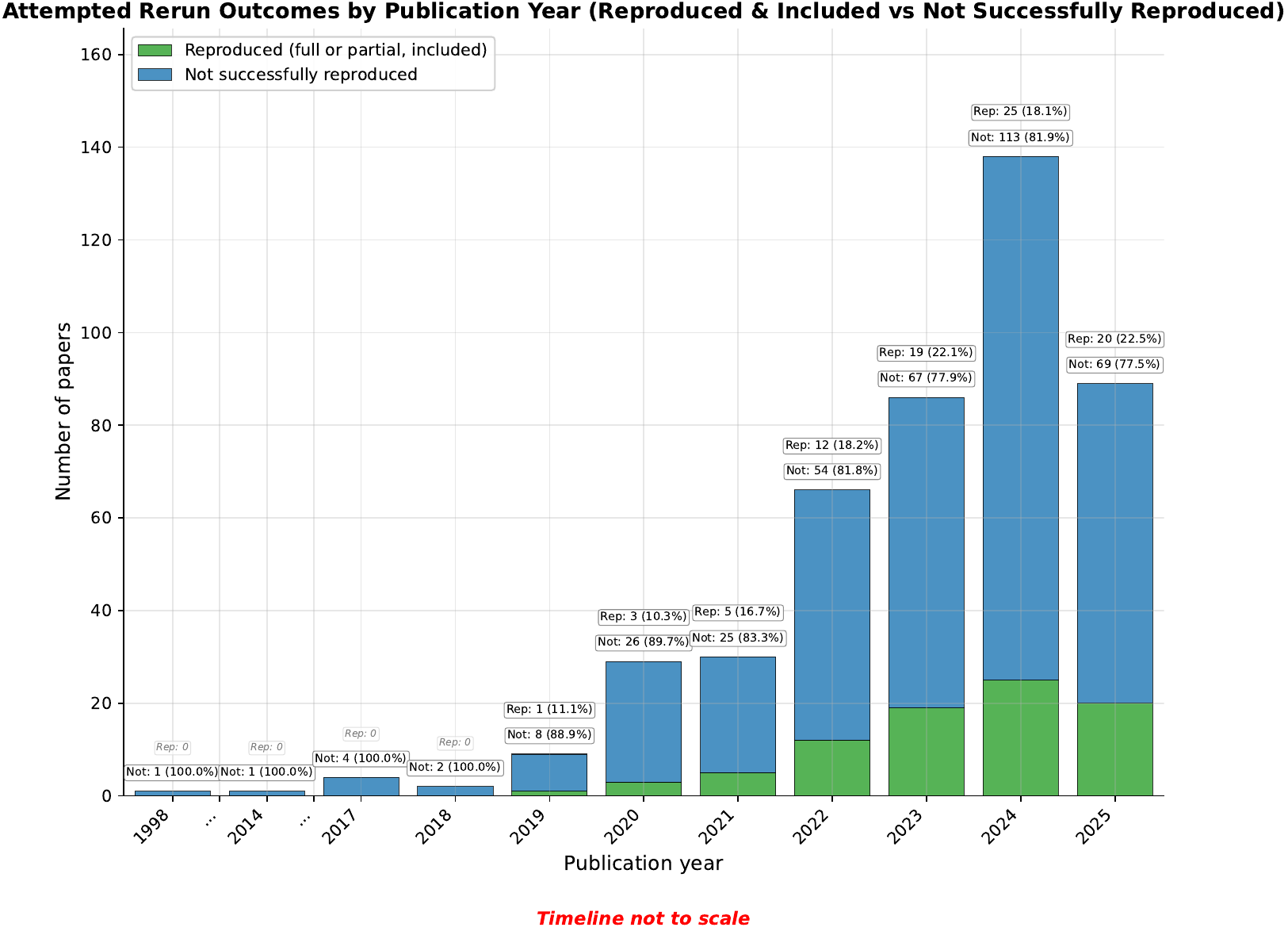}
    \caption{Reproduction outcomes over time for the attempted NSAI papers. Stacked bars show, for each publication year, the number of papers that were successfully reproduced (full or partial) versus not reproduced under the study protocol, illustrating the growth of the field alongside a reproduction rate that shows no sustained upward trend.}
    \label{fig:repro-by-year}
\end{figure*}

\begin{figure*}[t]
\centering
    \includegraphics[width=0.9\textwidth]{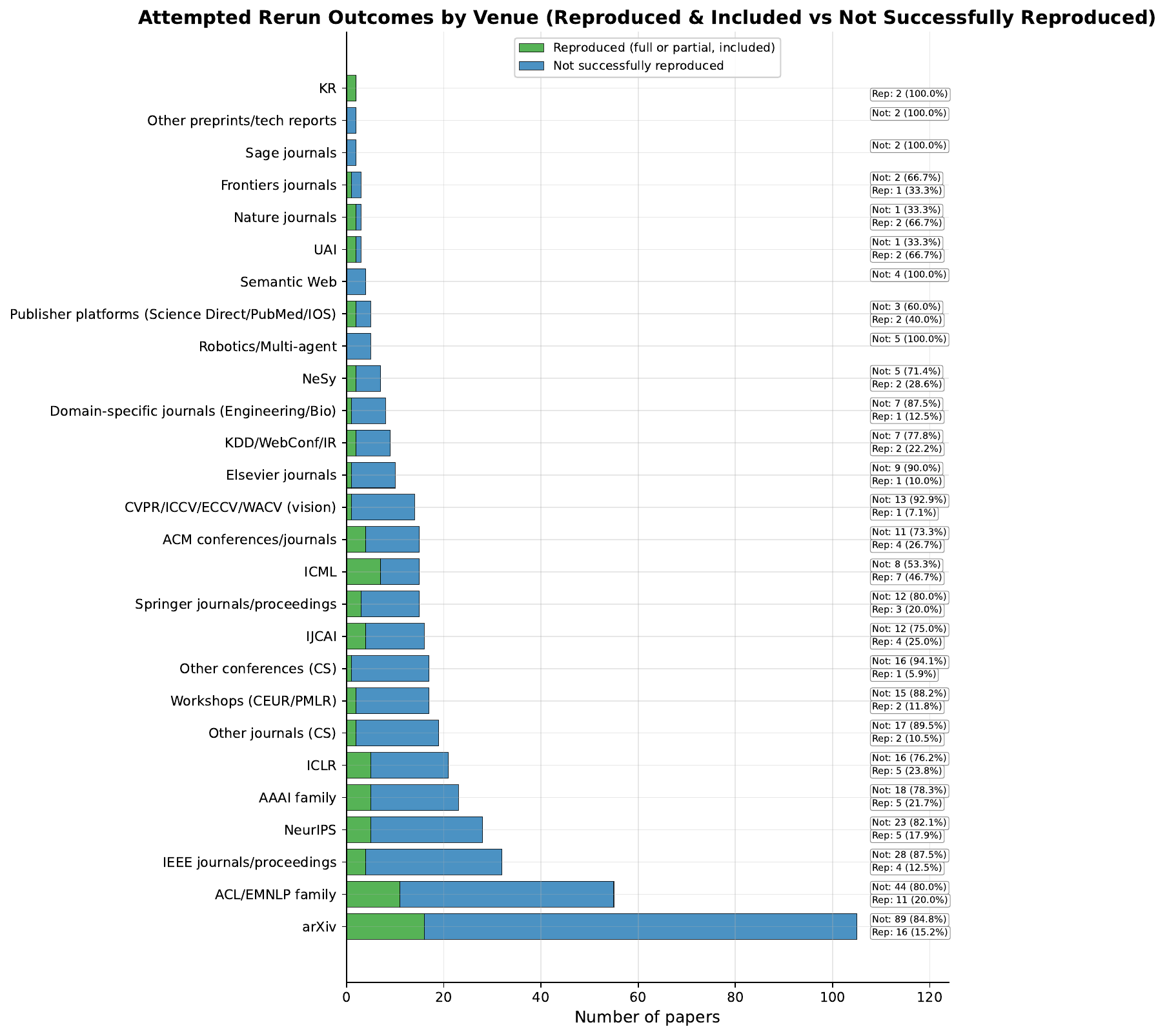}
    \caption{Reproduction outcomes by venue group for the attempted NSAI papers. Stacked bars show, for each venue family, the number of papers that were successfully reproduced (full or partial) versus not reproduced. Venues are ordered by number of attempted reruns.}
    \label{fig:repro-by-venue}
    \label{fig:repro-year-venue}
\end{figure*}
\else
\begin{figure}[t]
\centering
    \includegraphics[width=0.95\textwidth]{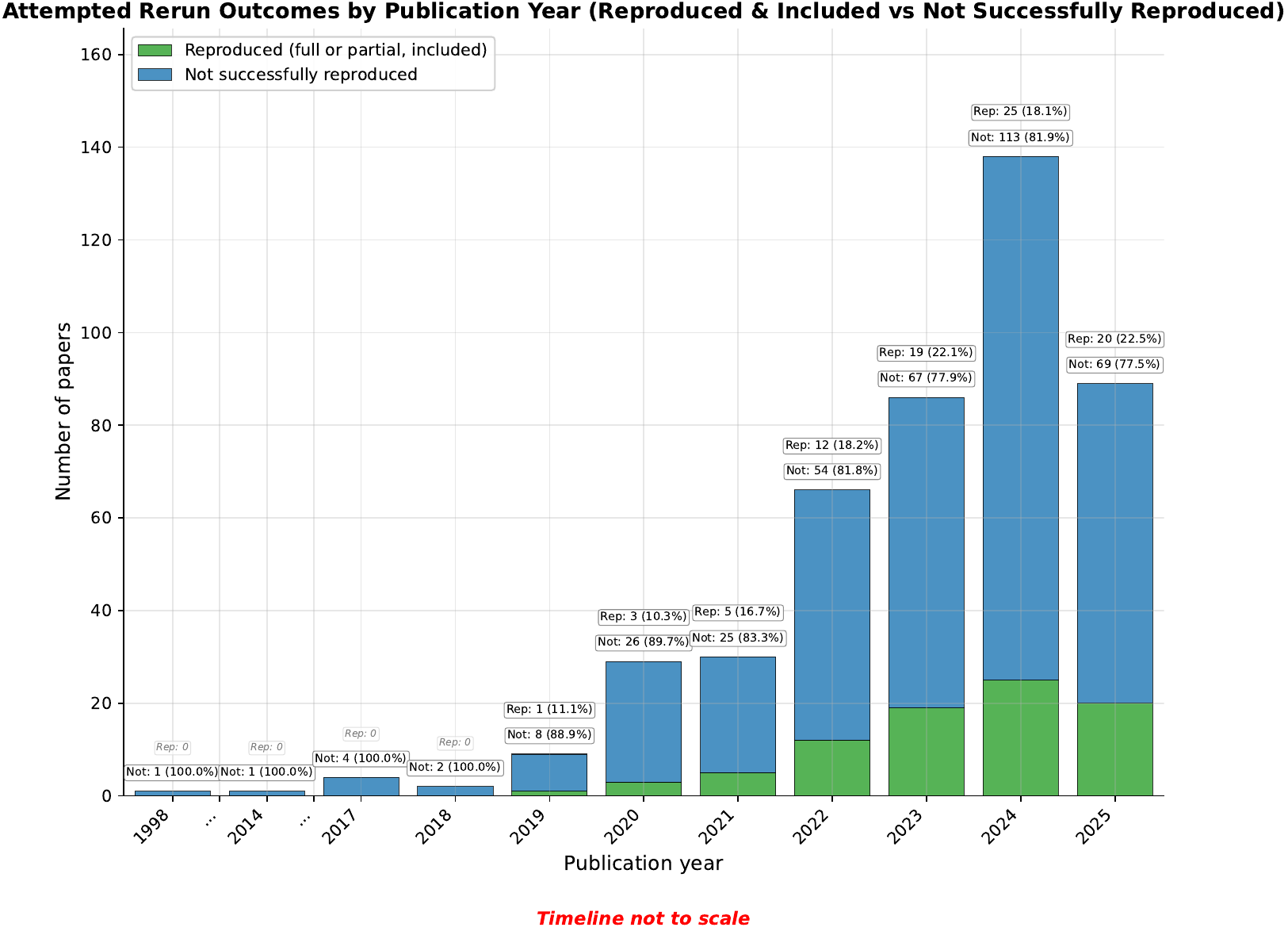}
    \caption{Reproduction outcomes over time for the attempted NSAI papers. Stacked bars show, for each publication year, the number of papers that were successfully reproduced (full or partial) versus not reproduced under the study protocol, illustrating the growth of the field alongside a reproduction rate that shows no sustained upward trend.}
    \label{fig:repro-by-year}
\end{figure}

\begin{figure}[t]
\centering
    \includegraphics[width=\textwidth]{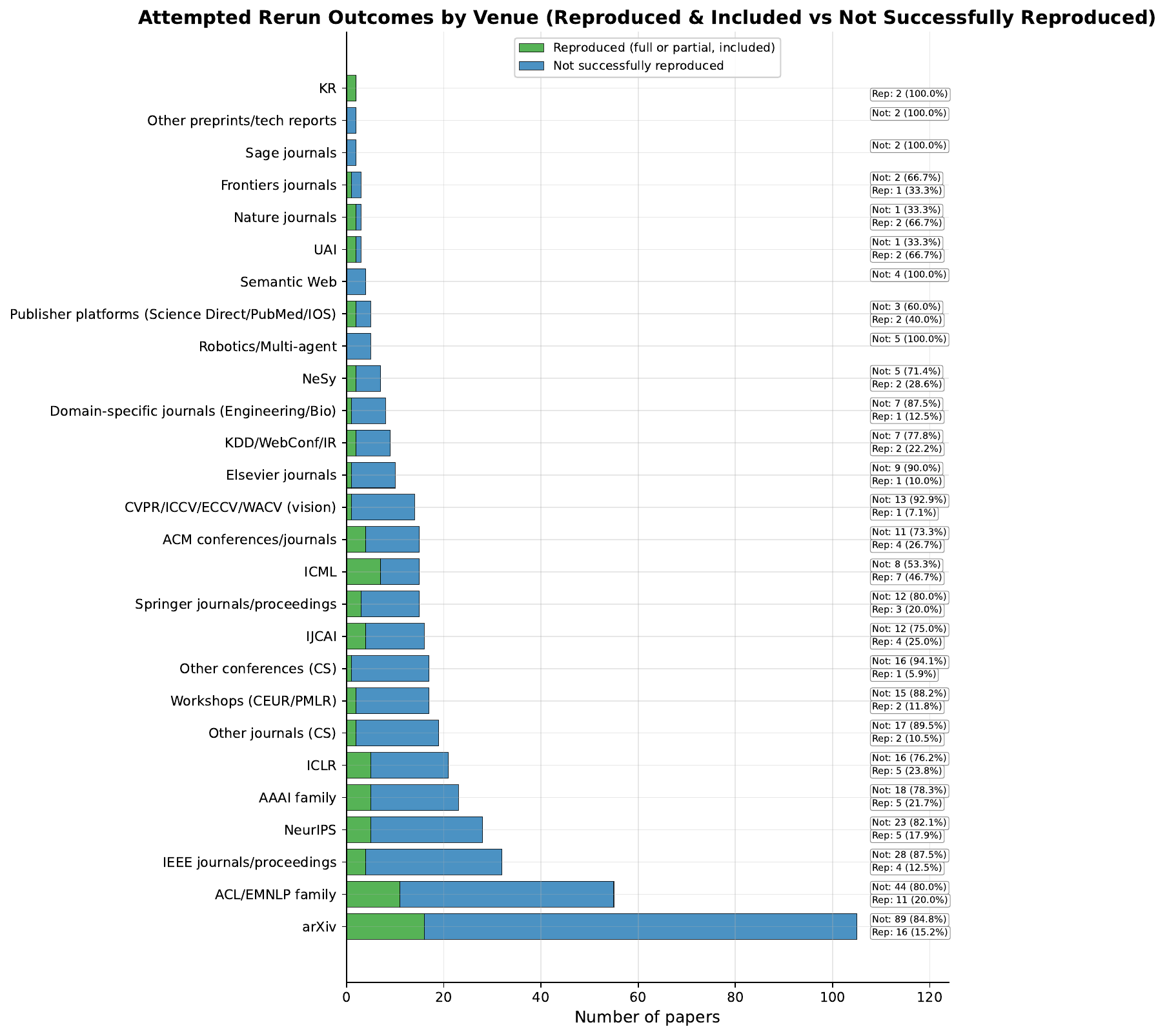}
    \caption{Reproduction outcomes by venue group for the attempted NSAI papers. Stacked bars show, for each venue family, the number of papers that were successfully reproduced (full or partial) versus not reproduced. Venues are ordered by number of attempted reruns.}
    \label{fig:repro-by-venue}
    \label{fig:repro-year-venue}
\end{figure}
\fi

\ifFullSizeMainFigures
\begin{figure*}[t]
\centering
\includegraphics[width=\textwidth]{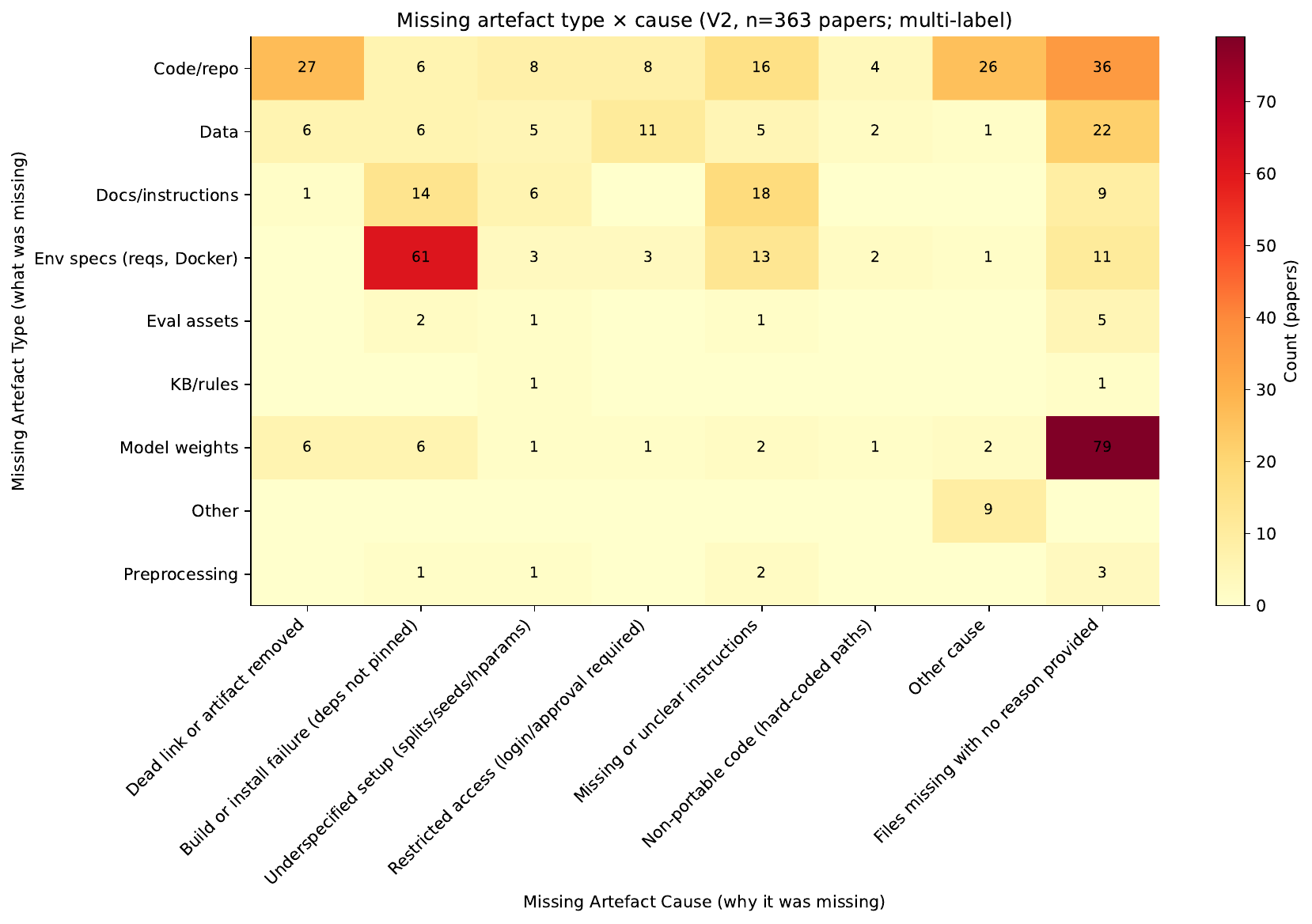}
  \caption{Rows indicate \emph{what} was missing and columns indicate \emph{why}. The plot contains non-exclusive, multi-label annotations across 363 attempted studies; a study may contribute to multiple cells, so cell counts do not sum to 363.}
  \label{fig:missing-type-cause-heatmap}
\end{figure*}
\else
\begin{figure}[t]
\centering
\includegraphics[width=\textwidth]{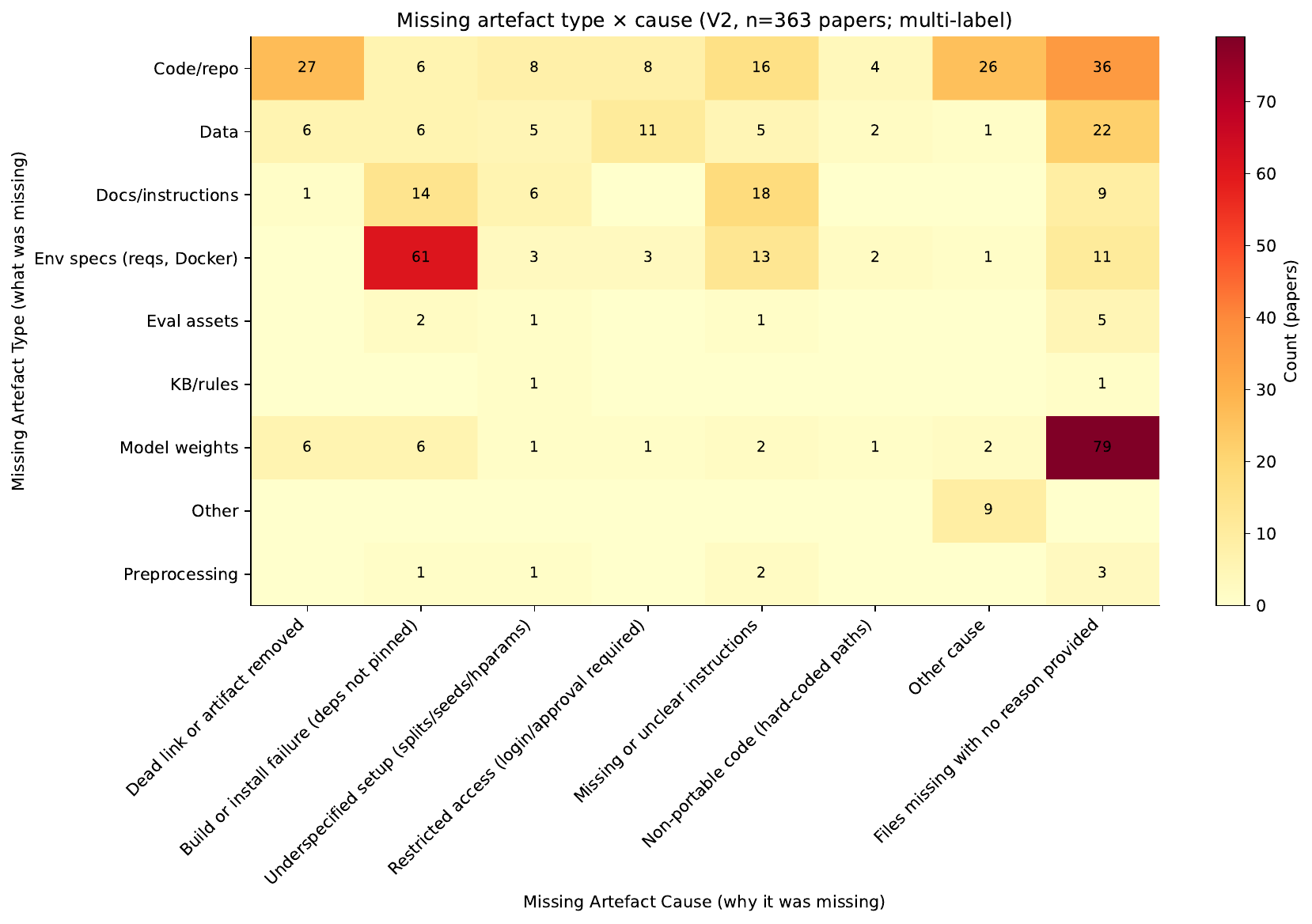}
  \caption{Rows indicate \emph{what} was missing and columns indicate \emph{why}. The plot contains non-exclusive, multi-label annotations across 363 attempted studies; a study may contribute to multiple cells, so cell counts do not sum to 363.}
  \label{fig:missing-type-cause-heatmap}
\end{figure}
\fi

\ifFullSizeMainFigures
\begin{figure*}[t]
\centering
\includegraphics[width=\textwidth]{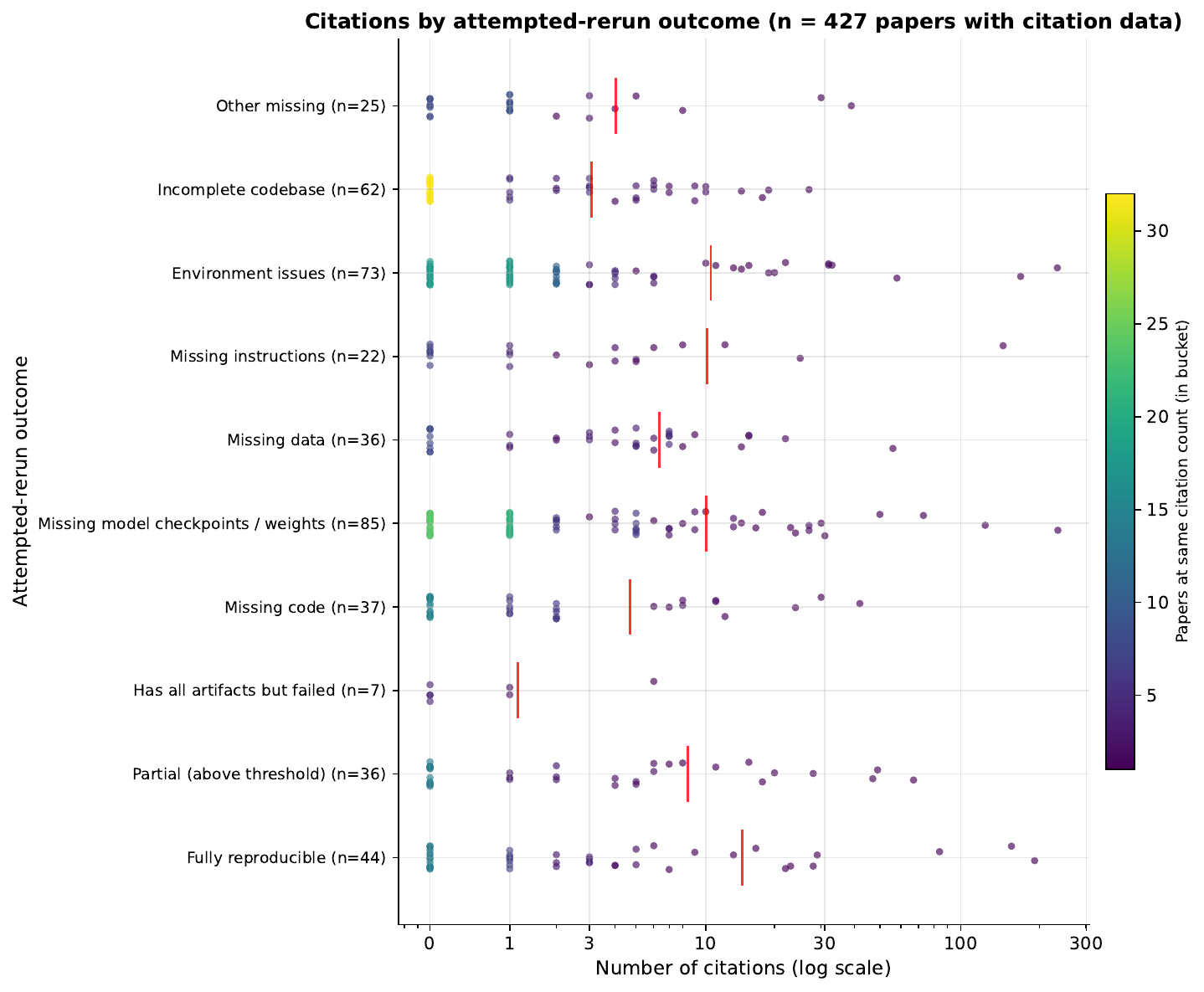}
  \caption{Citation counts by attempted-rerun outcome/failure reason for NSAI papers with available citation data (\(n=427\)) shown on a log scale. The red marker indicates the mean citation count for each bucket.}
  \label{fig:citations-by-outcome}
\end{figure*}
\else
\begin{figure}[t]
\centering
\includegraphics[width=\textwidth]{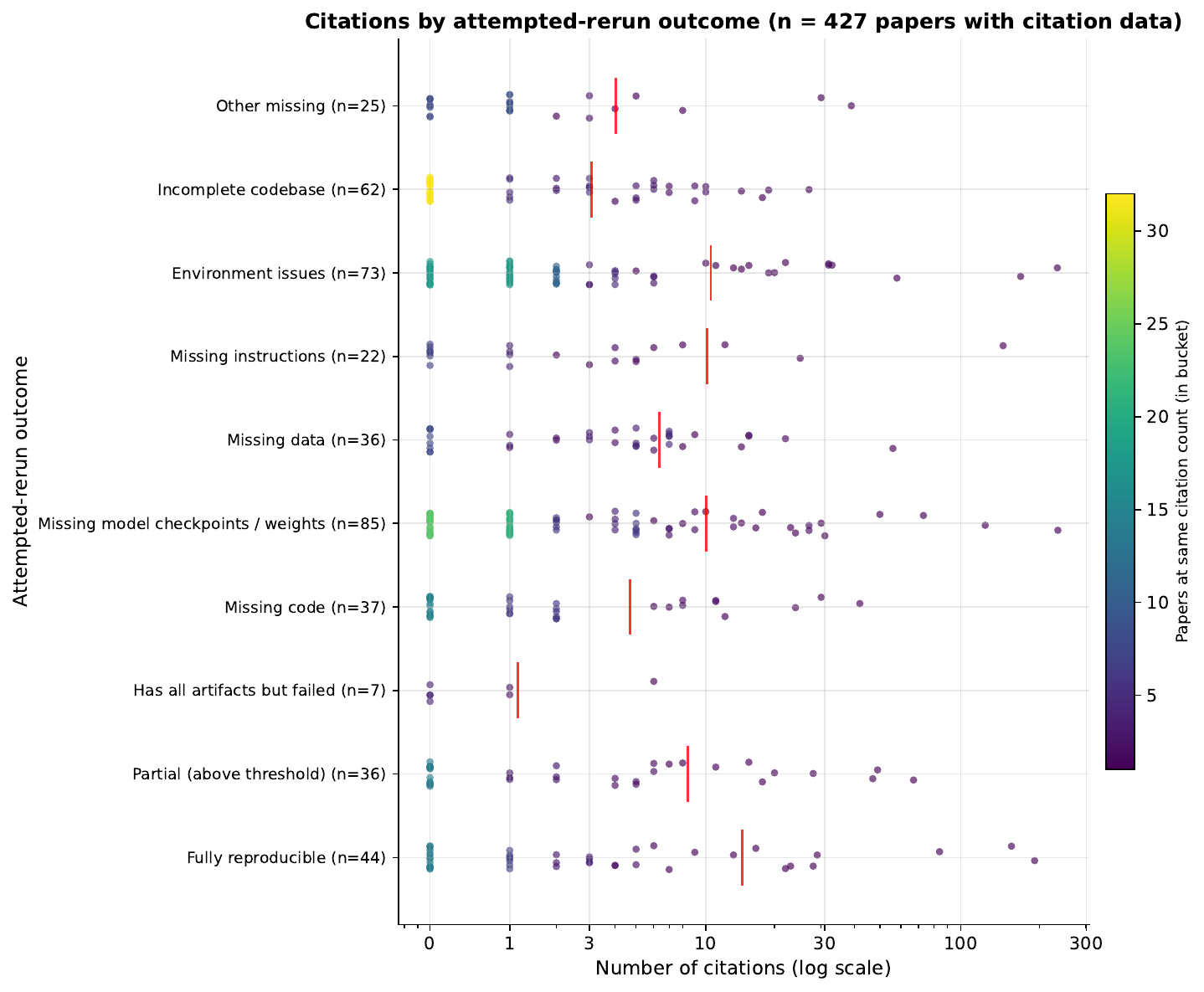}
  \caption{Citation counts by attempted-rerun outcome/failure reason for NSAI papers with available citation data (\(n=427\)) shown on a log scale. The red marker indicates the mean citation count for each bucket.}
  \label{fig:citations-by-outcome}
\end{figure}
\fi

\subsection{Six-Stage Audit Procedure}
\label{subsec:six-stage-procedure}
The operational procedure used the same six stages introduced above.

\noindent\textbf{Stage 1---Corpus construction and initial screening.}
The nine-database search was deduplicated and screened by title and
abstract, as described in Section~\ref{subsec:corpus-screening}.

\noindent\textbf{Stage 2---Full-text eligibility confirmation.}
Annotators rechecked language, topical relevance, original empirical
research, quantitative evaluation, and full-text accessibility. The
61 late full-text exclusions occurred at this stage.

\noindent\textbf{Stage 3---Repository identification and artifact inventory.}
Annotators located a public code artifact from the paper or structured
external search, recording its URL, license, and available commit hash.
Of the \num{1304} full-text-eligible papers, \num{849} did not pass this
public-code gate and \num{455} entered the rerun cohort. Within the rerun
cohort, the team inventoried code, data, model weights, evaluation assets,
environment specifications, and documentation. When access was gated,
up to two e-mail requests were made over 14 days; indispensable artifacts
remaining inaccessible were treated as post-entry failure outcomes.

\noindent\textbf{Stage 4---Environment build and executable integrity.}
Author-provided \texttt{Dockerfile}, \texttt{requirements.txt},
\texttt{environment.yml}, or equivalent specifications were used to
reconstruct the environment. Unit or smoke tests were run where possible.
Irreconcilable environment or code execution defects were evaluated under
$O_5$, while inaccessible indispensable inputs were evaluated under $O_4$.

\noindent\textbf{Stage 5---Result re-execution and outcome assignment.}
The primary experiment was attempted under the bounded-repair and
seven-day protocol described in Section~\ref{subsec:evaluation-design}.
For completed runs, annotators recorded reported versus reproduced metrics
and applied the stated fidelity criterion. The final study-level outcomes
used $O_1$--$O_5$ as defined above, distinguishing full reproduction,
partial reproduction, completed-but-unsuccessful reruns, artifact-access
blocks, and execution failures.

\noindent\textbf{Stage 6---Data extraction and outcome audit.}
For $O_1$ and $O_2$ papers, Sections 4--8 of the extraction form captured
neural and symbolic design, datasets, compute, metrics, and evaluation
procedures. One annotator subsequently reviewed all $O_1$/$O_2$ papers
and a stratified sample of $O_3$/$O_4$/$O_5$ records.

\subsection{Technical Infrastructure, Limitations, and Data Management}
\label{subsec:infra-limitations-data}

All reproductions were executed on \AuditCluster{}
under its default quality-of-service limits. Jobs without GPUs ran in the \texttt{standard} partition on up to a single 128-core Zen 3 node with 512 GiB RAM (7-day wall-clock cap). GPU jobs used the \texttt{GPU} partition, each limited to one full GPU node consisting of either 4 × NVIDIA A100 (40 GiB each, 128 Zen 3 CPU cores, 512 GiB RAM) or 4 × NVIDIA H100 (80 GiB each, 96 Sapphire Rapids cores, 512 GiB RAM), also with a 7-day limit. Software stacks were loaded per job via the cluster’s \texttt{module} system. Author-supplied Dockerfiles were converted to Apptainer images using \texttt{docker2singularity}, and when conversion failed, the build was completed on a local machine.

\subsubsection{Limitations}
The primary construct of this paper is reproducibility of reruns of same-artifacts to determine whether an independent team can obtain the reported result using the artifacts released by the authors. Under this construct, trained checkpoints were treated as indispensable when the published evaluation depended on a fixed trained model state. We did not substitute retraining in the primary analysis because retraining changes the experimental object and introduces additional stochastic, software, and platform variance. However, some papers without released checkpoints may remain reproducible in principle via retraining; we therefore interpret missing-checkpoint cases as non-rerunnable under this protocol rather than as proof that the underlying method is scientifically irreproducible in every broader sense. This framing is consistent with ACM’s current artifact terminology, under which reproducibility means an independent group obtaining the same result using the authors’ own artifacts, with agreement judged within an acceptable tolerance rather than exact identity.\footnote{\url{https://www.acm.org/publications/policies/artifact-review-and-badging-current}} Additionally, although annotators could patch environment files and author evaluation harnesses when absent from the original release, severe lack of documentation for code and environments, together with architecture-level defects intrinsic to the framework itself (for example, a fresh clone that crashes with tensor shape mismatches or NaN losses despite the original environment), remained failure conditions under the protocol.

\subsubsection{Data Management \& Transparency}
All repositories, weights, patched scripts, environment manifests, and execution logs required for reproduction of examined manuscripts remain on \AuditFileSystem{} under institutional retention policy.  Summary statistics are mirrored in the public extraction sheet\footnote{\ReproductionLog}. The complete audit trail for this systematic reproduction review (including per-paper reproduction logs, annotator tagging logs, and the reproduction protocol and supporting documentation) is archived in a long-term online repository \footnote{\AuditArchive}, together with the scripts required to regenerate every reported aggregate and figure. We additionally publish the raw outcomes via an interactive webpage\footnote{\InteractiveResultsBrowser} to enable paper-level inspection and independent recomputation of the headline counts reported in the manuscript.

\section{Results}
\subsection{System‑Level Reproducibility}
\label{sec:funnel}

Figure~\ref{fig:funnel-overall-pie} reports outcomes for the attempted rerun set only (\(n = 455\)). Within this set, \textbf{48} studies (\SI{10.55}{\percent}) were fully reproduced and \textbf{37} (\SI{8.13}{\percent}) were partially reproduced, yielding \textbf{85} successful reruns overall (\SI{18.68}{\percent}). Failures within the attempted set comprised \textbf{321} papers (\SI{70.55}{\percent}) assigned missing, incomplete, or environment-related artifact failures, \textbf{42} (\SI{9.23}{\percent}) blocked by unavailable or unusable code repositories, and \textbf{7} (\SI{1.54}{\percent}) that had the nominal artifact set but failed because of environment or code defects. The \textbf{61} records that failed late full-text eligibility checks are reported separately in Figure~\ref{fig:prisma} and are not counted as rerun outcomes. The rerun success rate is therefore $85/455 = 18.68\%$ (Wilson 95\% CI 15.4--22.5). Relative to the eligible corpus ($n = \num{1304}$), the rate is $85/1304 = 6.52\%$ (Wilson 95\% CI 5.3--8.0). The primary obstacle to successful same-artifact reruns was incomplete artifact availability. Within the attempted set, \textbf{321} papers were assigned missing, incomplete, or environment-related artifact failures, and \textbf{42} by missing or unusable code repositories. A residual \textbf{7} studies shared the nominal artifact set yet still failed because of environment or code defects. By contrast, within the group of studies recorded as having the required artifacts, reruns succeeded in \textbf{85} of \textbf{92} cases (92.4\,\%, Wilson 95\% CI 85.1--96.3). Missing artifacts concentrated in three components. Absent model checkpoints or weights blocked \textbf{89} attempts (19.56\,\% of attempted reruns), environment or specification problems blocked \textbf{79} (17.36\,\%), and incomplete codebases blocked \textbf{67} (14.73\,\%). These three categories account for \textbf{235} of \textbf{321} missing-artifact failures (\SI{73.2}{\percent}), leaving \textbf{86} failures distributed across missing datasets (\textbf{36}, \SI{7.91}{\percent}), missing or incomplete documentation (\textbf{25}, \SI{5.49}{\percent}), and other missing items (\textbf{25}, \SI{5.49}{\percent}). The skew implies that most failures arose from predictable artifact-release gaps.

\subsection{Year-Wise Reproducibility Patterns}
\label{sec:yearly-trends}

Figure\,\ref{fig:repro-by-year} tracks NSAI reproduction outcomes by publication year and shows a steep rise in paper volume, from single digits before 2019 to \num{138} publications in 2024 and \num{89} in the partial-year 2025 cohort. From 2019 onward the annual full-or-partial rate ranges from \SI{10.3}{\percent} (2020) to \SI{22.1}{\percent} (2023), with intermediate values of \SI{11.1}{\percent} (2019), \SI{16.7}{\percent} (2021), \SI{18.2}{\percent} (2022), and \SI{18.1}{\percent} (2024). Because the 2025 cohort covers only January--May, its denominator is still evolving; we therefore treat the \SI{22.5}{\percent} figure as provisional and exclude 2025 when describing the trend. The years before 2019 contributed \num{8} attempted reruns in total, none of which were reproduced, and we exclude them as well. Within the 2019 to 2024 window, the proportion of studies reproduced (fully or partially) remains in a narrow band of roughly \SI{10}{\percent} to \SI{22}{\percent} of the yearly output, and the stacked-bar profiles reveal no sustained upward trajectory.  In practical terms, the field is publishing substantially more work each year, but the likelihood that any given paper can be reproduced under our protocol has not materially improved, suggesting that the surge in publications has not, by itself, improved real-world reproducibility under our protocol.

\subsection{Venue-Level Impacts on Reproducibility}
\label{sec:venue}

Figure\,\ref{fig:repro-by-venue} compares reproduction outcomes across twenty-seven venue families and shows no detectable advantage for journals, conferences, or preprint
servers at the class level ($\chi^2 = 2.26$, $df = 2$, $p = .32$). Large outlets illustrate this point as arXiv achieves \textbf{16/105} successes (\SI{15.2}{\percent}), ACL/EMNLP family \textbf{11/55} (\SI{20.0}{\percent}), IEEE journals/proceedings \textbf{4/32} (\SI{12.5}{\percent}), and NeurIPS \textbf{5/28} (\SI{17.9}{\percent}). Comparable rates are observed for AAAI family journals/conferences (\SI{21.7}{\percent}) and for ICLR (\SI{23.8}{\percent}) despite differing review models, while domain-specific journals, vision conferences, and publisher platforms all cluster in the low-to-mid-teens. The only apparent outlier is ICML at \textbf{7/15} (\SI{46.7}{\percent}) however, its small denominator limits generality.  Aggregating by broad class, conferences reproduce at 21.1 \% (53/251, Wilson 95\% CI 16.5--26.6), journals at 16.5 \% (16/97, CI 10.4--25.1), and preprints at 15.0 \% (16/107, CI 9.4--22.9). The three intervals overlap substantially.  Publication venue alone is not a reliable predictor of NSAI reproducibility. Artifact completeness, not outlet type, is what distinguishes the attempts that succeeded from those that did not.

\subsection{Failure modes behind missing artifacts}
Figure~\ref{fig:missing-type-cause-heatmap} summarizes non-exclusive missing-artifact type and cause annotations for the $n=363$ attempted papers assigned artifact-related failure categories (321 missing/incomplete/environment-related and 42 missing/unusable code). A paper can contribute to more than one cell, so the cell counts do not sum to 363. The largest cell is missing model weights with files missing with no reason provided (79 papers), and the second is missing environment specifications linked to build or install failure from unpinned dependencies (61 papers). For model weights, restricted access and dead links are rare (1 and 6 papers, respectively), so the bottleneck is usually not permissioning but non-release or decay, and the claim that a repository enables rerunning the reported pipeline is falsified by inspection in the majority of weight-missing cases. Environment failures concentrate on dependency drift, which is consistent with repositories that omit a locked descriptor or ship one that no longer resolves under current tooling. Code and repository issues are also common, with files missing with no reason provided (36) and dead links or removed artifacts (27), plus a substantial remainder attributed to other causes (26), so \enquote{code available} is an unreliable proxy for an executable codebase. Data failures split between restricted access (11) and unexplained absence (22), while documentation failures cluster in missing or unclear instructions (18) and build failures (14).

\subsection{Citation impact vs reproducibility}
Figure \ref{fig:citations-by-outcome} demonstrates whether citation impact is a useful proxy for practical reproducibility by comparing DOI-resolved citation counts across the attempted-rerun outcome and failure-reason buckets summarized in Figures \ref{fig:funnel-overall-pie} and \ref{fig:missing-type-cause-heatmap} above. Citation counts were obtained by DOI lookup for 427 of the \num{455} attempted records. The remaining 28 lacked a resolvable DOI and are excluded here. The distribution of citations from the remaining 427 papers, as shown in Figure \ref{fig:citations-by-outcome}, indicates substantial overlap across all buckets, with a large mass of zero and low citation papers in every outcome and a long right tail that inflates the average number of citations per bucket. The fully reproducible bucket has the highest mean citation count at $14.1$ but a median of $2$, and comparably high means occur in dominant non-reproducible buckets such as missing model checkpoints or weights with mean $10.0$, median $1$, max $234$ and environment or specification issues with mean $10.5$, median $2$, max $233$, showing that incomplete artifact release shows no clear relationship with number of citations that a paper may receive.

\subsection{Answers to Research Questions RQ1 -- RQ3}

For \textbf{RQ1} [proportion of NSAI papers that can be reproduced], within the attempted rerun set ($n = 455$), 48 studies met the strict full-reproduction criterion (10.55 \%, Wilson 95\% CI 8.1--13.7) and a further 37 met the partial criterion (8.13 \%). The full-or-partial composite is 85/455 (18.68 \%). Relative to the full eligible corpus ($n = 1304$), these fall to 3.68\% (full; Wilson 95\% CI 2.8--4.8) and 6.52\% (full or partial; Wilson 95\% CI 5.3--8.0). For \textbf{RQ2} [effect of artifact completeness on reproduction success], artifact completeness is the dominant factor. Among 92 studies classified by the audit as having the required artifacts, 85 met the full-or-partial reproduction criterion (92.4\%). In contrast, 321 of the 455 attempts (70.55\%) were assigned missing, incomplete, or environment-related artifact failures, with missing model weights ($n = 89$), environment issues ($n = 79$), and incomplete codebases ($n = 67$) accounting for  73.2\% of those failures. For \textbf{RQ3} [variation in outcomes by publication year and venue family], the venue-class comparison was not statistically significant ($\chi^2=2.26$, $df=2$, $p=.32$). The descriptive annual full-or-partial success rates ranged from 10.3\% to 22.1\% between 2019 and 2024 without a sustained increase, and large outlets cluster in a similar narrow band, with arXiv at 15.2\%, ACL/EMNLP at  20.0\%, IEEE at 12.5\%, and NeurIPS at 17.9\%, and aggregate class rates of 21.1\% for conferences, 16.5\% for journals, and 15.0\% for preprints.

\section{Discussion}
\subsection{Summary of Principal Findings in Context}%
\label{subsec:principal_findings}

Each study in the attempted cohort ($n=455$) had a public code artifact that could be identified during screening. During the later audit, 42 of these repository links or code releases proved missing or unusable. In total, fewer than one in five attempted studies met the full-or-partial reproduction criterion. 

\subsection{Trends and Patterns}%
\label{subsec:trends_patterns}

Despite adopting an audit protocol that permits limited dependency updates, path corrections, and lightweight evaluation harnesses to be supplied by auditors, the proportion of NSAI papers that reproduced remained stubbornly flat, oscillating between \SI{10.3}{\percent} and \SI{22.1}{\percent} per year from 2019 through 2024. Figure~\ref{fig:funnel-overall-pie} attributes most failures to missing, incomplete, or environment-related artifact failures (\SI{70.55}{\percent}) and, to a lesser extent, to absent code (\SI{9.23}{\percent}), yet when the full range of artifacts required for reproduction is present, reproduction succeeds in \SI{92.4}{\percent} of cases.  Perhaps most striking is the insignificance of venues as a determinant for whether a manuscript will provide the resources required to reproduce the work. Conferences, journals, and preprints cluster at \SI{21.1}{\percent}, \SI{16.5}{\percent}, and \SI{15.0}{\percent} respectively, and the three confidence intervals overlap (Figure~\ref{fig:repro-by-venue}). The evidence points to artifact availability, rather than venue type or methodological novelty, as the main bottleneck to reproducibility in current NSAI research. The analysis, therefore, shifts the conversation from improving experimental technique to enforcing comprehensive artifact release across all publication outlets.

\subsection{\enquote{Dead} Codebase Links}
Despite passing the initial NSAI relevance screen and subsequent repository-identification step, 42 of the in-scope papers ultimately fell into the Missing Code category because the referenced repository was dead, private, empty, unrelated, or otherwise unusable at audit time. In most instances, the hyperlink supplied in the manuscript was (i) dead or resolving to a private or removed repository, (ii) redirected to a project homepage that described the system but held no source files, (iii) pointed to a repository that never held any source code, or (iv) led to code that was unrelated or insufficient to implement the reported method.  These false-positive disclosures illustrate that nominal compliance with \enquote{code available} guidelines is not enough, and that persistent, content-verified repositories and explicit version tags are essential if claims of openness are to translate into practical reproducibility.

\subsection{Reproducibility requirements for the domain of Neuro-Symbolic AI as a science}
\label{subsec:policy_future}

\subsubsection{Practical implications}  Systematic examination of the AAAI Author Reproducibility Checklist\footnote{\url{https://aaai.org/conference/aaai/aaai-26/reproducibility-checklist/}}, the NeurIPS Paper-Checklist Guidelines\footnote{\url{https://neurips.cc/public/guides/PaperChecklist}}, the IJCAI Reproducibility Guidelines\footnote{\url{https://www.ijcai.org/reproducibility}}, the NeurIPS report \cite{10.5555/3546258.3546422}, and the AI Magazine survey of reproducibility barriers and drivers \cite{10.1002/aaai.70002} shows that the five documents converge on a common baseline of artifacts that an empirical paper must provide: \emph{(i)} runnable source code; \emph{(ii)} the exact datasets or immutable links to them; \emph{(iii)} pre-trained checkpoints; \emph{(iv)} a single command or script that reproduces the reported metrics (an evaluation harness) \emph{(v)} a machine-readable environment file (\texttt{environment.yml}, \texttt{requirements.txt}, Dockerfile, etc.) and \emph{(vi)} documentation in the form of instructions that integrate items (i)–(v) into one executable workflow. All checklists emphasize persistent identifiers, explicit version locking, and permissive licensing.

\subsubsection{Policy levers}  
Existing venue initiatives fall into two categories. ACM artifact badges \footnote{\url{https://www.acm.org/publications/policies/artifact-review-and-badging-current}} include an external review of the repository, whereas the AAAI, NeurIPS, and IJCAI checklists are self-attested by the authors. Each submission to a peer-reviewed conference, journal, and even preprint venues should undergo an automated repository audit before peer review, confirming that all six artifacts are present and executable. Only 7 of 455 attempts were classified as failures despite nominally complete artifacts; an automated pre-review check could flag release deficiencies of the types observed in the other 363 unsuccessful attempts, although flagging does not itself resolve them.

\subsubsection{Future directions for NSAI Research}
Neuro‐Symbolic models integrate multiple components in complicated systems to deliver functional systems, so a single missing artifact can invalidate the entire pipeline. An absent checkpoint severs logic bindings, and undocumented rule sets could compromise evaluation and as such, authors must archive the complete artifact bundle (codebase, datasets, model weights, evaluation scripts, environment files and documentation) in a DOI-minting repository (e.g., Zenodo, HuggingFace-Hub, Figshare, OSF, or Mendeley Data).\footnote{These services issue persistent identifiers, maintain long-term storage, and provide metadata suitable for citation.} Computer science research is only science when its results withstand independent execution.  A manuscript and a transient GitHub link do not satisfy this condition.  Requiring the six-item artifact bundle, bound by immutable URIs for all publication venues (including pre-publication), is the most immediate, low-cost intervention available to the community. The good news is that the barrier to improvement is low. The included papers from this study demonstrate that full reproducibility is achievable at any publication venue when authors commit to a complete artifact bundle from the outset. The community can act now by depositing code,  data, model weights, environment files, and a single-command evaluation script in a DOI-minting repository such as Zenodo or HuggingFace-Hub at submission time, not as an afterthought. Venues can reinforce this by  requiring automated artifact checks before peer review begins, a low-cost intervention relative to the effort wasted reproducing or discarding 
non-reproducible work.

\subsection{Case Studies}\label{sec:case_studies}

\subsubsection{Case Study 1: Scallop-A Badge-Awarded Benchmark for Reproducibility}\label{sec:scallop-case}

The Scallop PLDI 2023 paper~\cite{rayyan-242085286} earned both the ACM \emph{Artifacts Available} and \emph{Artifacts Evaluated—Reusable} badges, and its release exemplifies full-stack reproducibility. A Zenodo snapshot (\href{https://zenodo.org/records/7804200}{DOI 7804200}) freezes the exact commit, dataset splits, pretrained weights, and SHA-256 hashes. Docker and Conda manifests plus a Rust nightly lockfile reconstitute the software stack, while a single script (\texttt{run\_all.sh}) rebuilds, trains, and evaluates the eight-task benchmark.  GitHub CI recompiles the core library, runs unit tests, and executes a smoke benchmark on every push, preventing configuration drift.  Our audit reproduced all eight tasks on a single A100 GPU with a median absolute deviation $\leq$ 3 \% from the reported metrics and required no manual intervention. Scallop, therefore, is a fantastic demonstration that badge-level artifact curation and automated validation can turn a complex NSAI system into a reliably re-runnable research object.

\subsubsection{Case Study~2: LogiCity-Conference-Level Reproducibility without Formal Badging}\label{sec:logicity-case}

Unlike Scallop, LogiCity~\cite{rayyan-242084181} earned no external badge, yet its NeurIPS 2024 Datasets and Benchmarks package reproduced on the first attempt.  The authors froze code, data, and checkpoints in a version-tagged release, shipped Docker and Conda manifests, and wired a one-command launcher that rebuilds the simulator, trains agents, and scores results. Continuous-integration runs the full smoke benchmark on every commit, and SHA-256 checks ensure downloads match the snapshot.  On our A100 test node, the Safe-Path-Following and Visual-Action-Prediction tasks reproduced within $\pm$ 2 pp of the reported scores and retained the original baseline ordering. LogiCity therefore demonstrates that rigorous curation and automated checks can deliver badge-level reproducibility at a premier conference—even when no formal badging program exists.

\subsubsection{Case Study 3: MARS-Preprint-Level Reproducibility through Curated artifacts}\label{sec:mars-case}

The \textit{Mechanism-of-Action Retrieval System} (MARS) \cite{rayyan-242084473} appears only as an arXiv preprint first posted in October 2024, yet its artifact package met every criterion for a gold standard reproducible package. A version-tagged GitHub repository provides Docker and Conda descriptors, immutable data archives, and pretrained checkpoints, and a single \texttt{run.sh} script rebuilds the biomedical-graph pipeline, trains the models, and computes evaluation metrics. Re-execution on one NVIDIA A100 GPU reproduced the MoA-Net benchmark with MRR 0.315 versus the reported 0.318 and Hits@10 0.672 versus 0.685 (absolute deviation $\leq$ 2 percentage points), preserving the published ranking of baselines. All code, data, and weights are protected by SHA-256 hashes, permitting table verification without retraining. Although the release has not yet undergone an external artifact audit, this example indicates that a fully versioned, one-command artifact stack can deliver robust reproducibility irrespective of publication venue.

\subsubsection{Cross-Case Comparison and Venue Implications}\label{sec:case-comparison}

The three case studies span the principal publication strata in computer science, as a badge-audited archival proceedings (Scallop), a top-tier conference track without external certification (LogiCity), and an open-access arXiv preprint (MARS), yet each enabled low-effort, reliable same-artifact reproduction by adhering to the same practical principles.  Each artifact package (i) freezes the software stack with Docker or Conda, (ii) archives immutable data and pretrained checkpoints under permanent identifiers, (iii) offers a one-command script that rebuilds and evaluates the full pipeline, (iv) discloses every configuration parameter, and (v) offers a good level of documentation with instructions on environment build requirements.  In the reported case-study results, Scallop had a median absolute deviation no greater than 3\%, the two named LogiCity tasks differed by at most two percentage points, and the reported MARS metrics differed by less than two percentage points. These summaries use the metric-specific deviation measures stated in the individual case studies. Scallop achieves this through a formal ACM badge review, LogiCity relies on conference-driven community norms and continuous-integration tests, and MARS depends on author-integrity. These results indicate that enforceable artifact standards, not venue prestige, are the decisive factor in achieving reliable reproducibility.

\section{Conclusion}
This study provides, to the best of our knowledge, the first large-scale, retrospective, publication-year-stratified assessment of same-artifact rerun reproducibility within neuro-symbolic AI.  Across \num{1\,365} self-identified NSAI records identified by our search, \num{1\,304} met full-text eligibility and \num{455} entered the attempted same-artifact rerun audit; \num{48} were fully reproduced and \num{37} were partially reproduced (\num{85} combined). Failure analysis shows that reproducibility typically collapses when essential artifacts such as full codebases, datasets, pretrained model weights, evaluation harness scripts, environmental files or documentation are absent at release. Computer science research attains scientific legitimacy only when independent investigators can verify its empirical claims.  To that end, we recommend that \emph{all} empirical submissions, regardless of venue, be required to pass an automated artifact audit \emph{prior} to peer review.  The audit should confirm the availability of six items including (i) full source code, (ii) immutable datasets, (iii) pretrained model checkpoints, (iv) an executable evaluation script, (v) a machine-readable environment descriptor, and (vi) concise documentation sufficient to invoke the pipeline. Persistent DOI-minting repositories such as Zenodo, HuggingFace-Hub, Figshare, OSF, or Mendeley Data already provide the necessary infrastructure to host these artifacts at scale and with minimal cost to authors. Adopting this minimal standard would align computer science publication practice with the norms of experimental science and render future advances in Neuro-Symbolic AI transparent, verifiable, and readily extensible.  The community must therefore institutionalize immutable artifact disclosure, requiring and archiving scientific works prior to publication, and thereby ensure that progress is measurable and cumulative by lowering the barrier for full pipeline reproduction of Neuro-Symbolic systems.

\bibliographystyle{IEEEtran}
\bibliography{references}

\begingroup
\setlength{\tabcolsep}{1pt}
\appendix
\section{Appendix}

\paragraph{Data Availability.} The supplementary materials supporting this study are deposited in the \AuditArchive. The archive contains the paper-level reproduction and data-extraction records, missing-artifact annotations, audit protocols, annotator documentation, analysis scripts, generated aggregate results, and an offline copy of the results browser. A worked example demonstrating application of the data-extraction form is available through the \DataExtractionDemo. Paper-level outcomes can also be inspected through the \InteractiveResultsBrowser.

\subsection{Broader impacts}
This work can positively affect the machine-learning community by encouraging more transparent, verifiable, and cumulative Neuro-Symbolic AI research. At the same time, stronger artifact-release requirements may increase burdens on authors, especially under-resourced groups, and may create pressure to release sensitive, proprietary, or legally restricted artifacts. These risks can be mitigated through anonymized review artifacts, persistent metadata, gated access, and documented exemptions where full public release is not ethically or legally possible.



%
%

\subsection{Final List of Reproduced Works}
A full list of the papers that were able to be reproduced for this study can be found here: \href{https://brandonio-c.github.io/NSAI-2025-Survey/}{NSAI-2025-reproducibility-study}.

\begin{onecolumn}
\begin{longtable}{|p{0.22\textwidth}|p{0.28\textwidth}|p{0.48\textwidth}|}
\hline
\textbf{Citation} & \textbf{Title} & \textbf{Description} \\
\hline
\endfirsthead

\hline
\textbf{Citation} & \textbf{Title} & \textbf{Description} \\
\hline
\endhead

\hline
\endfoot

\hline
\endlastfoot

    \small\raggedright Acharya et al. (2025)\\\small\href{https://github.com/lotussavy/IWCMC-2025/tree/main}{Codebase} & \href{https://www.researchgate.net/publication/388686587_Neurosymbolic_AI_for_Travel_Demand_Prediction_Integrating_Decision_Tree_Rules_into_Neural_Networks}{Neurosymbolic AI for Travel Demand Prediction: Integrating Decision Tree Rules into Neural Networks} & This paper introduces a neuro-symbolic AI framework that blends decision tree rules with neural networks to predict travel demand. By combining interpretability with deep learning power, it achieves more accurate and explainable results for transportation planning and resource optimization. \\
\hline
    \small\raggedright Ishay et al. (2023)\\\small\href{https://github.com/azreasoners/gpt-asp-rules?tab=readme-ov-file}{Codebase} & \href{https://doi.org/10.24963/kr.2023/37}{Leveraging Large Language Models to Generate Answer Set Programs} & They use LLMs to work through logic puzzle solving in a step-by-step manner. \\
\hline
    \small\raggedright Murali et al. (2019)\\\small\href{https://github.com/muraliadithya/vdp}{Codebase} & \href{https://arxiv.org/pdf/1907.05878}{Composing Neural Learning and Symbolic Reasoning with an Application to Visual Discrimination} & propose a compositional neurosymbolic framework that combines a neural network to detect objects and relationships with a symbolic learner that finds interpretable discriminators. \\
\hline
    \small\raggedright Ahmed et al. (2022)\\\small\href{https://github.com/KareemYousrii/SPL}{Codebase} & \href{https://proceedings.neurips.cc/paper_files/paper/2022/hash/c182ec594f38926b7fcb827635b9a8f4-Abstract-Conference.html}{Semantic probabilistic layers for neuro-symbolic learning} & The paper presents Semantic Probabilistic Layers (SPLs), a neural network module that ensures structured-output predictions are always consistent with logical constraints, enabling accurate and tractable neuro-symbolic learning. SPLs modularly combine probabilistic inference and logical reasoning, outperforming previous methods in tasks requiring strict output validity. \\
\hline
    \small\raggedright Ahmed et al. (2022)\\\small\href{https://github.com/UCLA-StarAI/NeSyEntropy}{Codebase} & \href{https://proceedings.mlr.press/v180/ahmed22a.html}{Neuro-symbolic entropy regularization} & This paper presents neuro-symbolic entropy regularization, a unified framework that combines entropy regularization and neuro-symbolic learning for structured prediction tasks. By constraining entropy minimization to outputs that form valid structures (as defined by logical circuits), the approach yields models that are both more accurate and more likely to produce valid predictions, demonstrated across semi-supervised and fully-supervised experiments. \\
\hline
    \small\raggedright Akl (2024)\\\small\href{https://github.com/HannaAbiAkl/NeSy-Code-Generation-Workflow}{Codebase} & \href{https://doi.org/10.48550/arXiv.2402.16910}{NeSy is alive and well: A LLM-driven symbolic approach for better code comment data generation and classification} & This paper presents a neuro-symbolic workflow combining semantic rule–based decomposition with an LLM to generate controlled synthetic data for C-code comment classification. Empirical results show that this augmentation improves ML models (Voting Classifier, Random Forest, MLP) performance measured in F1 score. \\
\hline
    \small\raggedright Alam et al. (2024)\\\small\href{https://github.com/FutureComputing4AI/Hadamard-derived-Linear-Binding}{Codebase} & \href{https://arxiv.org/abs/2410.22669}{A Walsh Hadamard Derived Linear Vector Symbolic Architecture} & The paper introduces the Hadamard-derived Linear Binding (HLB), a novel vector symbolic architecture that leverages properties of the Walsh-Hadamard transform for efficient, numerically stable vector binding in neuro-symbolic AI. HLB achieves state-of-the-art performance on both classical VSA benchmarks and selected deep learning tasks, outperforming previous VSA methods in terms of computational complexity, accuracy, and differentiability for modern neural architectures. \\
\hline
    \small\raggedright Daniele and Luciano (2022)\\\small\href{https://github.com/DanieleAlessandro/KENN2}{Codebase} & \href{https://doi.org/10.48550/arXiv.2205.15762}{Knowledge Enhanced Neural Networks for relational domains} & Paper extends knowledge enhanced NNs to handle relational data and shows that stacking multiple KE layers deals with rule dependencies, achieving better accuracy than baseline NNS on Citeseer citation network classification while also being faster than baseline Semantic Based Regularization (SBR) and Relational Neural Machines (RNM) \\
\hline
    \small\raggedright Ibrahimzada et al. (2024)\\\small\href{https://github.com/Intelligent-CAT-Lab/AlphaTrans}{Codebase} & \href{https://arxiv.org/pdf/2410.24117}{AlphaTrans: A Neuro-Symbolic Compositional Approach for Repository-Level Code Translation and Validation} & Solves the task of code translation from one programming language to another. It does so using neural symbolic framework that breaks down the source code into fragments and also utilized the test code to ensure code was properly translated. \\
\hline
    \small\raggedright Alon et al. (2022)\\\small\href{https://github.com/neulab/retomaton}{Codebase} & \href{http://proceedings.mlr.press/v162/alon22a.html}{Neuro-symbolic language modeling with automaton-augmented retrieval} & The paper presents RETOMATON, a neuro-symbolic system that approximates costly nearest-neighbor datastore searches in retrieval-based language models using automaton states and pointer links, enabling substantial speed-ups while maintaining or improving perplexity. \\
\hline
    \small\raggedright Asai and Muise (2020)\\\small\href{https://github.com/guicho271828/latplan}{Codebase} & \href{10.5555/3491440.3491811}{Learning neural-symbolic descriptive planning models via cube-space priors: The voyage home (to STRIPS)} & neuro-symbolic architecture is trained end-to-end to produce a succinct and effective discrete state transition model from images alone. \\
\hline
    \small\raggedright Aspis et al. (2022)\\\small\href{https://github.com/YanivAspis/Embed2Sym}{Codebase} & \href{https://scholar.archive.org/work/r3ff2msdz5gbdgatkevgg3dmem/access/wayback/https://proceedings.kr.org/2022/44/kr2022-0044-aspis-et-al.pdf}{Embed2sym-scalable neuro-symbolic reasoning via clustered embeddings} & The paper presents Embed2Sym, a scalable framework that combines neural perception and symbolic reasoning through clustered embeddings, enabling fast training, interpretability, and generalization in tasks that exceed the scalability of prior neuro-symbolic systems. Embed2Sym achieves state-of-the-art results and significantly reduces training time on complex reasoning tasks involving visual and symbolic inputs. \\
\hline
    \small\raggedright Badreddine et al. (2022)\\\small\href{https://github.com/logictensornetworks/logictensornetworks}{Codebase} & \href{https://doi.org/10.1016/j.artint.2021.103649}{Logic Tensor Networks} & This paper presents Logic Tensor Networks (LTN), a neurosymbolic framework that supports querying, learning and reasoning with both rich data and abstract knowledge about the world. LTN introduces a fully differentiable logical language, called Real Logic, whereby the elements of a first-order logic signature are grounded onto data using neural computational graphs and first-order fuzzy logic semantics \\
\hline
    \small\raggedright Bairi et al. (2024)\\\small\href{https://github.com/microsoft/codeplan}{Codebase} & \href{https://doi.org/10.1145/3643757}{CodePlan: Repository-Level Coding using LLMs and Planning} & The paper introduces CodePlan, a framework that treats large-scale repository-level code edits (e.g., API migrations or temporal edits across many files) as a planning problem: it uses an LLM guided by static dependency and impact analysis to generate a chain of edits until the repository satisfies a correctness oracle. They evaluate on C\# and Python codebases and show that it outperforms no-planning baselines in build success rate and matching ground truth edits. \\
\hline
    \small\raggedright Barbiero et al. (2023)\\\small\href{https://github.com/pietrobarbiero/pytorch_explain}{Codebase} & \href{http://proceedings.mlr.press/v202/barbiero23a.html}{Interpretable neural-symbolic concept reasoning} & The paper proposes the Deep Concept Reasoner (DCR), an interpretable concept-based model that uses neural networks to generate fuzzy logic rules from concept embeddings, and then executes those rules on concept truth degrees to make semantically meaningful and differentiable predictions. \\
\hline
    \small\raggedright Barnaby et al. (2023)\\\small\href{https://github.com/celestebarnaby/ImageEye}{Codebase} & \href{https://arxiv.org/pdf/2304.03253}{ImageEye: Batch Image Processing using Program Synthesis} & Batch editing of images such as cropping out a desired object/person in a batch of 100+ images has not been an easy task. The paper describes a neuro symbolic approach to generate programs from user demonstrations that perform such tasks. They show the program can automate 96\% of these tasks. \\
\hline
    \small\raggedright Biggio et al. (2021)\\\small\href{https://github.com/SymposiumOrganization/NeuralSymbolicRegressionThatScales}{Codebase} & \href{https://proceedings.mlr.press/v139/biggio21a.html}{Neural symbolic regression that scales} & The paper presents NeSymReS, a neural symbolic regression model that leverages large-scale pre-training of Transformers on procedurally generated equations and data, enabling scalable, efficient, and robust discovery of symbolic equations from input-output data pairs. The approach outperforms traditional and neural symbolic regression methods across diverse evaluation benchmarks. \\
\hline
    \small\raggedright Li et al. (2024)\\\small\href{https://github.com/Jaraxxus-Me/LogiCity}{Codebase} & \href{https://proceedings.neurips.cc/paper_files/paper/2024/file/8196be81e68289d7a9ece21ed7f5750a-Paper-Datasets_and_Benchmarks_Track.pdf}{LogiCity: Advancing Neuro-Symbolic AI with Abstract Urban Simulation} & LogiCity models diverse urban elements using semantic and spatial concepts, such as IsAmbulance(X) and IsClose(X, Y). These concepts are used to define FOL rules that govern the behavior of various agents. Since the concepts and rules are abstractions, they can be universally applied to cities with any agent compositions, facilitating the instantiation of diverse scenarios \\
\hline
    \small\raggedright Carraro et al. (2024)\\\small\href{https://github.com/tommasocarraro/NESYKnowledgeTransfer}{Codebase} & \href{https://link.springer.com/chapter/10.1007/978-3-031-56063-7_15}{Mitigating data sparsity via neuro-symbolic knowledge transfer} & Paper uses Logic Tensor Networks (LTN) to mitigate data sparsity issues in recommender systems. It combines Matrix Factorization (neural) models with First-Order Logic axioms (symbolic). \\
\hline
    \small\raggedright Chanin and Hunter (2023)\\\small\href{https://github.com/chanind/amr-social-chemistry-reasoner}{Codebase} & \href{https://arxiv.org/pdf/2303.08264}{Neuro-symbolic commonsense social reasoning} & present a novel system for taking social rules of thumb (ROTs) in natural language from the Social Chemistry 101 dataset and converting them to first-order logic where reasoning is performed using a neuro-symbolic theorem prover. \\
\hline
    \small\raggedright Dickens et al. (2024)\\\small\href{https://github.com/linqs/dickens-icml24}{Codebase} & \href{https://arxiv.org/abs/2401.09651}{Convex and Bilevel Optimization for Neuro-Symbolic Inference and Learning} & The paper presents a bilevel, gradient-based optimization framework for neural-symbolic (NeSy) learning, introducing a dual block coordinate descent algorithm and smooth formulation for efficient and scalable parameter learning, empirically validated on multiple datasets. \\
\hline
    \small\raggedright Chen and Jabbarvand (2024)\\\small\href{https://github.com/Intelligent-CAT-Lab/FlakyDoctor}{Codebase} & \href{https://doi.org/10.1145/3650212.3680369}{Neurosymbolic Repair of Test Flakiness} & Introduces FlakyDoctor, a neuro-symbolic repair system that couples LLM-based patch synthesis with static analysis, compilation “stitching,” and targeted validation to repair order-dependent (OD) and implementation-dependent (ID) flaky tests. Evaluated on 873 flaky tests from 243 projects, achieving \textasciitilde{}57\% OD and 59\% ID repair success and producing 79 previously un-fixed patches (19 merged PRs) \\
\hline
    \small\raggedright Chen et al. (2022)\\\small\href{https://github.com/czyssrs/ConvFinQA?tab=readme-ov-file}{Codebase} & \href{https://arxiv.org/pdf/2210.03849}{ConvFinQA: Exploring the Chain of Numerical Reasoning in Conversational Finance Question Answering} & Introduces a new dataset to study chain of numerical reasoning in question-answering. \\
\hline
    \small\raggedright Glanois et al. (2021)\\\small\href{https://github.com/claireaoi/hierarchical-rule-induction}{Codebase} & \href{https://arxiv.org/abs/2112.13418}{Neuro-Symbolic Hierarchical Rule Induction} & This paper proposes an efficient and interpretable neuro-symbolic model called Hierarchical Rule Induction (HRI) to solve Inductive Logic Programming (ILP) problems. This model is built from a set of meta-rules organized in a hierarchical structure, and it invents first-order rules by learning embeddings to match facts and predicates. \\
\hline
    \small\raggedright Yin et al. (2024)\\\small\href{https://github.com/lemonsis/MDD-5k}{Codebase} & \href{https://doi.org/10.1609/aaai.v39i24.34763}{MDD-5k: A New Diagnostic Conversation Dataset for Mental Disorders Synthesized via Neuro-Symbolic LLM Agents} & Introduces a new framework to generate a dataset of diagnostic conversations between a patient and a doctor. It also provides a dataset with 5k high-quality long conversations with diagnosis results and treatment options. \\
\hline
    \small\raggedright Pryor et al. (2022)\\\small\href{https://github.com/linqs/neupsl-ijcai23}{Codebase} & \href{https://dl.acm.org/doi/10.24963/ijcai.2023/461}{NeuPSL: Neural Probabilistic Soft Logic} & Introduce Neural Probabilistic Soft Logic (NeuPSL), a novel neurosymbolic (NeSy) framework that unites state-of-the-art symbolic reasoning with the low-level perception of deep neural networks. To model the boundary between neural and symbolic representations, they propose a family of energy-based models, NeSy Energy-Based Models, and show that they are general enough to include NeuPSL and many other NeSy approaches. \\
\hline
    \small\raggedright Dagan et al. (2023)\\\small\href{https://github.com/itl-ed/llm-dp}{Codebase} & \href{https://arxiv.org/abs/2308.06391}{Dynamic Planning with a LLM} & presents LLM Dynamic Planner (LLM-DP): a neuro-symbolic framework where an LLM works hand-in-hand with a traditional planner to solve an embodied task. Given action-descriptions, LLM-DP solves Alfworld faster and more efficiently than a naive LLM ReAct baseline. \\
\hline
    \small\raggedright Daggitt et al. (2024)\\\small\href{https://github.com/vehicle-lang/vehicle}{Codebase} & \href{https://arxiv.org/abs/2401.06379}{Vehicle: Bridging the embedding gap in the verification of neuro-symbolic programs} & This paper presents Vehicle, a tool designed to bridge the "embedding gap" in the verification of neuro-symbolic programs—programs combining neural (machine learning) and symbolic (traditional code) components. Vehicle provides a unified, dependently-typed language and compiler that enables formal specifications for neural components to be integrated across different verification and training backends, demonstrated by verifying a neural network controller for an autonomous car. \\
\hline
    \small\raggedright Li et al. (2022)\\\small\href{https://github.com/danyangl6/nn-tli}{Codebase} & \href{https://ieeexplore-ieee-org.proxy-um.researchport.umd.edu/stamp/stamp.jsp?tp=&arnumber=10156357}{Learning Signal Temporal Logic through Neural Network for Interpretable Classification} & design a novel time function and sparse softmax function to improve the soundness and precision of the neural-STL framework. As a result, we can efficiently learn a compact STL formula for the classification of time-series data through off-theshelf gradient-based tools. \\
\hline
    \small\raggedright DeLong et al. (2024)\\\small\href{https://github.com/laurendelong21/MARS?tab=readme-ov-file#replicates}{Codebase} & \href{https://doi.org/10.48550/arXiv.2410.05289}{Mars: A neurosymbolic approach for interpretable drug discovery} & Mechanism of Action Retrieval System (MARS) is a neurosymbolic approach for drug discovery that uses logical rules with learned rule weights. The model has better interpretability through dynamic weight learning from logical rules. \\
\hline
    \small\raggedright Dhanraj and Eliasmith (2025)\\\small\href{https://github.com/vdhanraj/Neurosymbolic-LLM}{Codebase} & \href{https://aclanthology.org/2025.emnlp-main.1556.pdf}{Improving Rule-based Reasoning in LLMs via Neurosymbolic Representations} & introduces a novel neurosymbolic method that improves LLM reasoning by encoding hidden states into neurosymbolic vectors, enabling problem-solving within a neurosymbolic vector space \\
\hline
    \small\raggedright Done et al. (2023)\\\small\href{https://github.com/dong-river/CoRRPUS}{Codebase} & \href{https://doi.org/10.18653/v1/2023.findings-acl.832}{CORRPUS: Code-based Structured Prompting for Neurosymbolic Story Understanding} & CoRRPUS shows the usefuleness of code-based symbolic representations for enabling LLMs to perofrm better on story reasoning tasks. \\
\hline
    \small\raggedright Dong et al. (2019)\\\small\href{https://github.com/google/neural-logic-machines}{Codebase} & \href{https://arxiv.org/pdf/1904.11694}{Neural Logic Machines} & propose the Neural Logic Machine (NLM), a neural-symbolic architecture for both inductive learning and logic reasoning. NLMs exploit the power of both neural networks—as function approximators, and logic programming—as a symbolic processor for objects with properties, relations, logic connectives, and quantifiers. \\
\hline
    \small\raggedright Eiter et al. (2023)\\\small\href{https://github.com/pudumagico/NSGRAPH}{Codebase} & \href{https://ceur-ws.org/Vol-3432/paper11.pdf}{A Modular Neurosymbolic Approach for Visual Graph Question Answering} & This paper presents a modular neuro-symbolic architecture that processes images of graph-structured data (rather than symbolic graphs) by first using optical graph recognition and OCR to extract nodes/edges and labels, then parsing the question and using answer-set programming (ASP) for logical reasoning. This paper also introduces a new VGQA task and dataset (in 3 sets: tiny, small, medium) and establishes a baseline of 73\% accuracy. \\
\hline
    \small\raggedright Misino et al. (2022)\\\small\href{https://github.com/EleMisi/VAEL}{Codebase} & \href{https://dl.acm.org/doi/10.5555/3600270.3600607}{VAEL: Bridging Variational Autoencoders and Probabilistic Logic Programming} & This paper presents VAEL, a neuro-symbolic generative model integrating variational autoencoders (VAE) with the reasoning capabilities of probabilistic logic (L) programming. Besides standard latent subsymbolic variables, their model exploits a probabilistic logic program to define a further structured representation, which is used for logical reasoning. \\
\hline
    \small\raggedright Marconato et al. (2024)\\\small\href{https://github.com/samuelebortolotti/bears}{Codebase} & \href{https://dl.acm.org/doi/10.5555/3702676.3702790}{BEARS Make Neuro-Symbolic Models Aware of their Reasoning Shortcuts} & They propose to ensure NeSy models are aware of the semantic ambiguity of the concepts they learn, thus enabling their users to identify and distrust low-quality concepts. Starting from three simple desiderata, they derive bears (BE Aware of Reasoning Shortcuts), an ensembling technique that calibrates the model's concept-level confidence without compromising prediction accuracy, thus encouraging NeSy architectures to be uncertain about concepts affected by RSs. \\
\hline
    \small\raggedright van Krieken et al. (2025)\\\small\href{https://github.com/HEmile/neurosymbolic-diffusion}{Codebase} & \href{https://arxiv.org/pdf/2505.13138}{Neurosymbolic Diffusion Models} & To overcome the limitations of the independence assumption, this paper introduces neurosymbolic diffusion models (NESYDMS), a new class of NeSy predictors that use discrete diffusion to model dependencies between symbols. their approach reuses the independence assumption from NeSy predictors at each step of the diffusion process, enabling scalable learning while capturing symbol dependencies and uncertainty quantification. \\
\hline
    \small\raggedright van Krieken et al. (2022)\\\small\href{https://github.com/HEmile/a-nesi.git}{Codebase} & \href{https://arxiv.org/abs/2212.12393}{A-NeSI: A Scalable Approximate Method for Probabilistic Neurosymbolic Inference} & Introduce Approximate Neurosymbolic Inference (A-NESI): a new framework for PNL that uses neural networks for scalable approximate inference. A-NESI 1) performs approximate inference in polynomial time without changing the semantics of probabilistic logics; 2) is trained using data generated by the background knowledge; 3) can generate symbolic explanations of predictions; and 4) can guarantee the satisfaction of logical constraints at test time, which is vital in safety-critical applications. \\
\hline
    \small\raggedright Endo et al. (2023)\\\small\href{https://github.com/markendo/HumanMotionQA/tree/master/NSPose}{Codebase} & \href{https://arxiv.org/abs/2305.08953}{Motion Question Answering via Modular Motion Programs} & Proposes a neuro symbolic framework to reason about motion sequences in human actions via symbolic reasoning and modular design. They say that it grounds motion through learning motion concepts, attribute neural operators and temporal relations. The task they chose is human motion QA for evaluate their NSPose method. \\
\hline
    \small\raggedright Karabulut et al. (2025)\\\small\href{https://arxiv.org/pdf/2504.19354}{Codebase} & \href{https://arxiv.org/abs/2504.19354}{Neurosymbolic Association Rule Mining from Tabular Data} & Aerial+ trains an autoencoder on tabular data, then extracts association rules by feeding in test vectors with features set to 1. If the autoencoder reconstructs other features with high probability, it creates a rule. Generates way fewer rules than FP-Growth (2-10x less) with better coverage and runs faster on big datasets, plus makes rule-based classifiers like CORELS train faster without losing accuracy. \\
\hline
    \small\raggedright Evans et al. (2021)\\\small\href{https://github.com/RichardEvans/apperception}{Codebase} & \url{https://www.sciencedirect.com/science/article/pii/S0004370221000722?via%3Dihub}{Making sense of raw input} & The central contribution of this paper is a neuro-symbolic framework for distilling interpretable theories out of streams of raw, unprocessed sensory experience. \\
\hline
    \small\raggedright Feinman and Lake (2020)\\\small\href{https://github.com/rfeinman/GNS-Modeling}{Codebase} & \href{https://arxiv.org/abs/2006.14448}{Learning task-general representations with generative neuro-symbolic modeling} & develop a generative neuro-symbolic (GNS) model of handwritten character concepts that uses the control flow of a probabilistic program, coupled with symbolic stroke primitives and a symbolic image renderer, to represent the causal and compositional processes by which characters are formed. \\
\hline
    \small\raggedright Feng et al. (2022)\\\small\href{https://github.com/feng-yufei/NS-NLI?tab=readme-ov-file}{Codebase} & \href{10.1162/tacl_a_00458}{Neuro-symbolic natural logic with introspective revision for natural language inference} & Presents a NeSy framework to integrate natural logic with NNs to do natural language inference. They use RL to help guide this natural logic. Overall this improves the intuitive inference understandability of humans. \\
\hline
    \small\raggedright Gong et al. (2024)\\\small\href{https://github.com/NanxuGong/feature-selection-via-autoregreesive-generation}{Codebase} & \href{https://arxiv.org/abs/2404.17157}{Neuro-Symbolic Embedding for Short and Effective Feature Selection via Autoregressive Generation} & Proposes a neuro-symbolic framework that models feature selection as an autoregressive token generation problem, learning embeddings for short and effective feature subsets with superior predictive performance compared to classical baselines. \\
\hline
    \small\raggedright Hakim et al. (2025)\\\small\href{https://github.com/sbhakim/ansr-dt}{Codebase} & \href{https://arxiv.org/abs/2501.08561}{ANSR-DT: An Adaptive Neuro-Symbolic Learning and Reasoning Framework for Digital Twins} & ANSR-DT presents an adaptive neuro-symbolic framework integrating deep learning (CNN-LSTM) and symbolic reasoning (Prolog) for digital twins. The system generates interpretable insights, learns rules from data, and provides knowledge graph visualizations to support decision-making in industrial scenarios. \\
\hline
    \small\raggedright Princis et al. (2024)\\\small\href{https://github.com/henrijsprincis/Xander}{Codebase} & \href{https://arxiv.org/abs/2408.13888}{Enhancing SQL Query Generation with Neurosymbolic Reasoning} & Find a way to use language models to generate SQL queries. They introduce a new tool called Xander which helps \\
\hline
    \small\raggedright Ho Fung et al. (2024)\\\small\href{https://github.com/hftsoi/SymbolNet}{Codebase} & \href{https://arxiv.org/abs/2401.09949}{SymbolNet: Neural Symbolic Regression with Adaptive Dynamic Pruning} & SymbolNet combines neural networks with symbolic regression through adaptive pruning to extract interpretable mathematical expressions while achieving high model compression. The method uses a custom training loop with threshold-based pruning to identify and eliminate less important network connections, enabling recovery of human-readable symbolic formulas from trained models without sacrificing predictive accuracy. \\
\hline
    \small\raggedright Howard et al. (2023)\\\small\href{https://github.com/IntelLabs/multimodal_cognitive_ai/tree/main/NeuroComparatives}{Codebase} & \href{https://arxiv.org/abs/2305.04978}{Neurocomparatives: Neuro-symbolic distillation of comparative knowledge} & This paper presents NeuroComparatives, a dataset of 8.8M comparative commonsense statements (e.g., "cats are typically smaller than dogs") generated using constrained beam search with GPT-2 and filtered using a discriminator trained on human annotations. The dataset aims to provide high-validity comparative knowledge for commonsense reasoning tasks. \\
\hline
    \small\raggedright Hu et al. (2023)\\\small\href{https://github.com/ant-research/StructuredLM_RTDT}{Codebase} & \href{https://arxiv.org/pdf/2303.02860}{A multi-grained self-interpretable symbolic-neural model for single/multi-labeled text classification} & propose a Symbolic-Neural model that can learn to explicitly predict class labels of text spans from a constituency tree without requiring any access to spanlevel gold labels \\
\hline
    \small\raggedright Ishay et al. (2024)\\\small\href{https://github.com/azreasoners/CRCG}{Codebase} & \href{https://arxiv.org/pdf/2506.10753}{Think before You Simulate: Symbolic Reasoning to Orchestrate Neural Computation for Counterfactual Question Answering} & The paper discusses a novel approach to do causal and temporal reasoning in video particularly for counterfactual reasoning. They make use of a causal graph and ASP to coordinate between perception and simulation modules. \\
\hline
    \small\raggedright Maene et al. (2024)\\\small\href{https://github.com/ML-KULeuven/klay?tab=readme-ov-file}{Codebase} & \href{https://arxiv.org/abs/2410.11415}{KLay: Accelerating Arithmetic Circuits for Neurosymbolic AI} & The paper presents KLay, a scalable data structure and layerization algorithm that accelerates arithmetic circuit computation in neurosymbolic AI systems, enabling efficient forward and backward passes. It demonstrates near-linear runtime scaling across five orders of magnitude, significantly outperforming baseline methods on logic-based inference tasks. \\
\hline
    \small\raggedright Ying et al. (2025)\\\small\href{https://github.com/Jie0618/PhysicsRegression}{Codebase} & \href{https://arxiv.org/abs/2503.07994}{A Neural Symbolic Model for Space Physics} & PhyE2E presents an end-to-end transformer-based framework for physics-informed symbolic regression that incorporates dimensional analysis as an inductive bias through oracle-guided divide-and-conquer strategies combined with MCTS refinement. The method achieves state-of-the-art performance on the Feynman equations benchmark, improving symbolic accuracy by 10-29\% over existing methods like PySR, uDSR, TPSR, and LaSR. \\
\hline
    \small\raggedright Kricheli et al. (2024)\\\small\href{https://github.com/lab-v2/PyEDCR}{Codebase} & \href{https://arxiv.org/pdf/2407.15192}{Error Detection and Constraint Recovery in Hierarchical Multi-Label Classification without Prior Knowledge} & his paper presents an approach that uses Error Detection Rules (EDR) to learn the failure modes of machine learning models, relaxing the common assumption that hierarchical constraints must exist beforehand. These learned rules are effective at detecting classifier errors and can be leveraged as constraints for Hierarchical Multi-label Classification (HMC), allowing for the recovery of explainable constraints even when they are not provided initially. \\
\hline
    \small\raggedright Ma et al. (2020)\\\small\href{https://github.com/Mayer123/HyKAS-CSKG}{Codebase} & \href{https://ojs.aaai.org/index.php/AAAI/article/view/17593/17400}{Knowledge-driven Data Construction for Zero-shot Evaluation in Commonsense Question Answering} & vary the set of language models, training regimes, knowledge sources, and data generation strategies, and measure their impact across tasks. Extending on prior work, we devise and compare four constrained distractor-sampling strategies \\
\hline
    \small\raggedright Katz et al. (2021)\\\small\href{https://github.com/garrettkatz/poppy-muffin}{Codebase} & \href{https://doi.org/10.3389/fnbot.2021.744031}{Tunable Neural Encoding of a Symbolic Robotic Manipulation Algorithm} & present a neurocomputational controller for robotic manipulation based on the recently developed “neural virtual machine” (NVM). Authors program the NVM with a symbolic algorithm that solves blocks-world restacking problems, and execute it in a robotic simulation environment \\
\hline
    \small\raggedright Kelly et al. (2023)\\\small\href{https://github.com/alex-calderwood/there-and-back}{Codebase} & \href{https://ojs.aaai.org/index.php/AIIDE/article/view/27502}{There and back again: extracting formal domains for controllable neurosymbolic story authoring} & Demonstrate that languagemodels can be used to author narrative planning domainsfrom natural language stories with minimal human intervention. Second, authors explore the reverse, demonstrating that we can use logical story domains and plans to produce storiesthat respect the narrative commitments of the planne \\
\hline
    \small\raggedright Kohaut et al. (2024)\\\small\href{https://github.com/HRI-EU/ProMis}{Codebase} & \href{10.1109/TITS.2025.3609835}{Probabilistic Mission Design in Neuro-Symbolic Systems} & Authors propose Probabilistic Mission Design (ProMis), a novel neuro-symbolic approach to navigating UAS within legal frameworks. ProMis is an interpretable and adaptable system architecture that links uncertain geospatial data and noisy perception with declarative, Hybrid Probabilistic Logic Programs (HPLP) to reason over the agent’s state space and its legality \\
\hline
    \small\raggedright Kohaut et al. (2024)\\\small\href{https://github.com/HRI-EU/ProMis/tree/cofi}{Codebase} & \href{10.48550/arXiv.2412.18347}{The Constitutional Filter} & introduces an approach for Bayesian estimation of agents expected to comply with a human-interpretable neuro-symbolic model we call its Constitution. Hence, autors present the Constitutional Filter (CoFi), leading to improved tracking of agents by leveraging expert knowledge, incorporating deep learning architectures, and accounting for environmental uncertainties \\
\hline
    \small\raggedright Kouvaros and Botoeva (2024)\\\small\href{https://github.com/NeuralMAS/venmas}{Codebase} & \href{https://ceur-ws.org/Vol-3819/short1.pdf}{Formal verification of parameterised neural-symbolic multi-agent systems} & This paper presents VENMAS, a formal verification toolkit for multi-agent systems with neural components, enabling bounded temporal logic verification of parameterized neural-symbolic agents. The work extends existing verification methods to handle systems with an arbitrary number of homogeneous agents using symbolic abstraction techniques. \\
\hline
    \small\raggedright Kulmanov et al. (2021)\\\small\href{https://github.com/bio-ontology-research-group/machine-learning-with-ontologies}{Codebase} & \href{https://doi.org/10.1093/bib/bbaa199}{Semantic similarity and machine learning with ontologies} & provide an overview over the methods that use ontologies to compute similarity and incorporate them in machine learning methods; in particular, the authors outline how semantic similarity measures and ontology embeddings can exploit the background knowledge in ontologies and how ontologies can provide constraints that improve machine learning models. \\
\hline
    \small\raggedright Kulmanov et al. (2024)\\\small\href{https://github.com/bio-ontology-research-group/deepgo2}{Codebase} & \href{https://www.nature.com/articles/s42256-024-00795-w}{Protein function prediction as approximate semantic entailment} & The Gene Ontology (GO) is a formal, axiomatic theory with over 100,000 axioms that describe the molecular functions, biological processes and cellular locations of proteins in three subontologies. Developed DeepGO-SE, a method that predicts GO functions from protein sequences using a pretrained large language model. DeepGO-SE generates multiple approximate models of GO, and a neural network predicts the truth values of statements about protein functions in these approximate models. \\
\hline
    \small\raggedright Lee et al. (2024)\\\small\href{https://github.com/THU-KEG/DiaKoP}{Codebase} & \href{https://dl.acm.org/doi/pdf/10.1145/3627673.3679229}{Diakop: Dialogue-based knowledge-oriented programming for neural-symbolic knowledge base question answering} & present Dialogue-based Knowledge-oriented Programming system (DiaKoP), a system with a chat interface designed for multi-turn knowledge base question answering (KBQA). DiaKoP enables users to decompose complex questions into multiple simpler follow-up questions and interact with the system to obtain answers. \\
\hline
    \small\raggedright Li et al. (2020)\\\small\href{https://github.com/liqing-ustc/NGS}{Codebase} & \href{https://proceedings.mlr.press/v119/li20f.html}{Closed loop neural-symbolic learning via integrating neural perception, grammar parsing, and symbolic reasoning} & This paper proposes a Neural-Grammar-Symbolic (NGS) model that improves upon inefficient reinforcement learning approaches by using a grammar model to bridge neural perception and symbolic reasoning, along with a novel back-search algorithm to learn from incorrect predictions. The experiments, conducted on handwritten formula recognition and visual question answering tasks, demonstrate that this method significantly outperforms reinforcement learning models in terms of performance, convergence speed, and data efficiency. \\
\hline
    \small\raggedright Li et al. (2022)\\\small\href{https://github.com/nju-websoft/AdaLoGN}{Codebase} & \href{https://aclanthology.org/2022.acl-long.494.pdf}{AdaLoGN: Adaptive Logic Graph Network for Reasoning-Based Machine Reading Comprehension} & present a neural-symbolic approach which, to predict an answer, passes messages over a graph representing logical relations between text units. \\
\hline
    \small\raggedright Li et al. (2023)\\\small\href{https://github.com/scallop-lang/scallop}{Codebase} & \href{https://dl.acm.org/doi/10.1145/3591280}{Scallop: A language for neurosymbolic programming} & present Scallop, a language which combines the benefits of deep learning and logical reasoning. Scallop enables users to write a wide range of neurosymbolic applications and train them in a data- and compute-efficient manner. It achieves these goals through three key features: 1) a flexible symbolic representation that is based on the relational data model; 2) a declarative logic programming language that is based on Datalog and supports recursion, aggregation, and negation; and 3) a framework for automatic and efficient differentiable reasoning that is based on the theory of provenance semirings. \\
\hline
    \small\raggedright Li et al. (2024)\\\small\href{https://github.com/scallop-lang/scallop}{Codebase} & \href{https://doi.org/10.1609/aaai.v38i9.28934}{Relational Programming with Foundation Models} & implement VIEIRA by extending the SCALLOP compiler with a foreign interface that supports foundation models as plugins. We implement plugins for 12 foundation models including GPT, CLIP, and SAM. We evaluate VIEIRA on 9 challenging tasks that span language, vision, and structured and vector databases. Our evaluation shows that programs in VIEIRA are concise, can incorporate modern foundation models, and have comparable or better accuracy than competitive baselines. \\
\hline
    \small\raggedright Murphy et al. (2024)\\\small\href{https://github.com/loganrjmurphy/LeanEuclid}{Codebase} & \href{https://dl.acm.org/doi/10.5555/3692070.3693567}{Autoformalizing Euclidean Geometry} & use theorem provers to fill in such diagrammatic information automatically, so that the LLM only needs to autoformalize the explicit textual steps, making it easier for the model. \\
\hline
    \small\raggedright Manginas et al. (2024)\\\small\href{https://github.com/nmanginas/nesya}{Codebase} & \href{https://www.ijcai.org/proceedings/2025/0662.pdf}{NeSyA: Neurosymbolic Automata} & how that symbolic automata can be integrated with neural-based perception, under probabilistic semantics towards an end-to-end differentiable model. Their proposed hybrid model, termed NESYA (Neuro Symbolic Automata) is shown to either scale or perform more accurately than previous NeSy systems in a synthetic benchmark and to provide benefits in terms of generalization compared to purely neural systems in a realworld event recognition task \\
\hline
    \small\raggedright Mejri et al. (2024)\\\small\href{https://github.com/mmejri3/LARS-VSA}{Codebase} & \href{https://www.researchgate.net/publication/380820165_LARS-VSA_A_Vector_Symbolic_Architecture_For_Learning_with_Abstract_Rules}{LARS-VSA: A Vector Symbolic Architecture For Learning with Abstract Rules} & The paper presents LARS-VSA, a neuro-symbolic framework leveraging hyperdimensional computing for abstract rule learning with compositional generalization. It introduces a high-dimensional attention mechanism and demonstrates superior generalization, efficiency, and accuracy over contemporary neural and neuro-symbolic baselines on multiple relational reasoning and math tasks. \\
\hline
    \small\raggedright Olausson et al. (2023)\\\small\href{https://github.com/benlipkin/linc}{Codebase} & \href{https://aclanthology.org/2023.emnlp-main.313.pdf}{(sic) LINC: A Neurosymbolic Approach for Logical Reasoning by Combining Language Models with First-Order Logic Provers} & investigate the validity of instead reformulating such tasks as modular neurosymbolic programming, which we call LINC: Logical Inference via Neurosymbolic Computation. In LINC, the LLM acts as a semantic parser, translating premises and conclusions from natural language to expressions in first-order logic. These expressions are then offloaded to an external theorem prover, which symbolically performs deductive inference. \\
\hline
    \small\raggedright Post et al. (2024)\\\small\href{https://github.com/clairepost/AMRtoUMR}{Codebase} & \href{https://aclanthology.org/2024.dmr-1.15/}{Accelerating UMR adoption: Neuro-symbolic conversion from AMR-to-UMR with low supervision} & The paper proposes a neuro-symbolic method for converting AMR (Abstract Meaning Representation) roles to UMR (Uniform Meaning Representation) roles by integrating animacy parsing and logic rules with a neural network. The approach addresses non-deterministic role mappings with minimal human supervision, achieving 75.81\% accuracy compared to a 62.35\% baseline neural network. \\
\hline
    \small\raggedright Premsri and Kordjamshidi (2024)\\\small\href{https://github.com/HLR/SpaRTUNQChain}{Codebase} & \href{https://arxiv.org/abs/2406.13828}{Neuro-symbolic Training for Reasoning over Spatial Language} & This paper presents SpaRTUNQChain, a neuro-symbolic framework that combines BERT embeddings with spatial logic constraints for question-answering over spatial language. The system uses the DomiKnowS framework to enforce compositional reasoning rules during training, improving accuracy on spatial reasoning benchmarks. \\
\hline
    \small\raggedright Quan et al. (2025)\\\small\href{https://github.com/neuro-symbolic-ai/peirce}{Codebase} & \href{https://aclanthology.org/2025.acl-demo.2.pdf}{Peirce: Unifying material and formal reasoning via llm-driven neuro-symbolic refinement} & introduce PEIRCE, a neuro-symbolic framework designed to unify material and formal inference through an iterative conjecture–criticism process. \\
\hline
    \small\raggedright Roig Vilamala et al. (2023)\\\small\href{https://github.com/MarcRoigVilamala/DeepProbCEP}{Codebase} & \href{https://doi.org/10.1016/j.eswa.2022.119376}{DeepProbCEP: A neuro-symbolic approach for complex event processing in adversarial settings} & Introduces DeepProbCEP, a differentiable ProbLog + CNN hybrid that trains end‑to‑end from complex‑event labels. \\
\hline
    \small\raggedright Skryagin et al. (2024)\\\small\href{https://github.com/ml-research/SLASH}{Codebase} & \href{https://jair.org/index.php/jair/article/view/15027}{Scalable Neural-Probabilistic Answer Set Programming} & This paper introduces Answer Set Networks (ASNs), a novel neural-symbolic solver based on Graph Neural Networks (GNNs) designed to overcome the high computational costs and CPU-bound nature of traditional Answer Set Programming (ASP) solvers. By translating ASP problems into a graph format that leverages GPU parallelization, ASNs outperform existing systems and enable new applications like fine-tuning Large Language Models (LLMs) with logic and solving large-scale drone navigation tasks. \\
\hline
    \small\raggedright Tammet et al. (2023)\\\small\href{https://github.com/tammet/nlpsolver}{Codebase} & \href{https://link.springer.com/chapter/10.1007/978-3-031-38499-8_29}{An Experimental Pipeline for Automated Reasoning in Natural Language (Short Paper)} & The paper presents NLPSolver, an experimental end-to-end pipeline that parses English text with a neural UD parser, converts it into extended first-order logic with defaults and confidences, performs reasoning with a high-performance default logic engine, and converts proofs back into natural language answers and explanations. The authors evaluate the system on toy NLI/QA examples, a subset of HANS, and AllenAI ProofWriter demos, showing strong performance there and highlighting the pipeline as a basis for future hybrid neuro-symbolic systems. \\
\hline
    \small\raggedright Winters et al. (2021)\\\small\href{https://github.com/ML-KULeuven/deepstochlog}{Codebase} & \href{https://arxiv.org/pdf/2106.12574}{DeepStochLog: Neural Stochastic Logic Programming} & This paper introduces DeepStochLog, a novel neural-symbolic framework that integrates neural networks into stochastic definite clause grammars (SDCGs) to define a probability distribution over possible derivations. The experimental evaluation shows this approach scales significantly better than methods based on neural probabilistic logic programs and achieves state-of-the-art results on several challenging neural-symbolic tasks. \\
\hline
    \small\raggedright Trinh et al. (2024)\\\small\href{https://github.com/google-deepmind/alphageometry}{Codebase} & \href{https://www.nature.com/articles/s41586-023-06747-5}{Solving olympiad geometry without human demonstrations} & AlphaGeometry is a neuro-symbolic system that uses a neural language model, trained from scratch on our large-scale synthetic data, to guide a symbolic deduction engine through infinite branching points in challenging problems. \\
\hline
    \small\raggedright Wu et al. (2024)\\\small\href{https://github.com/SoftWiser-group/LaM4Inv}{Codebase} & \href{https://dl-acm-org.proxy-um.researchport.umd.edu/doi/pdf/10.1145/3691620.3695014}{LLM Meets Bounded Model Checking: Neuro-symbolic Loop Invariant Inference} & proposes LaM4Inv, a neuro-symbolic framework that combines large language models (Llama-3-8B, GPT-3.5, GPT-4, GPT-4-Turbo) with bounded model checking (ESBMC + SMT solvers) to automatically infer loop invariants for C programs. It evaluates LaM4Inv on an expanded benchmark of 316 loop-invariant problems, showing large gains in the number of verified loops compared to classical invariant generators and recent LLM-based baselines. \\
\hline
    \small\raggedright Wu and Nakayama (2024)\\\small\href{https://ieeexplore.ieee.org/abstract/document/10680497}{Codebase} & \href{https://ieeexplore.ieee.org/abstract/document/10680497/}{MILE: Memory-Interactive Learning Engine for Neuro-Symbolic Solutions to Mathematical Problems} & MILE is a neuro-symbolic math word problem solver that predicts formulas using a memory-based decoder instead of tree-structured decoding, letting it handle more complex computation graphs. It beats existing methods on Math23K by learning formulas as rules. \\
\hline
    \small\raggedright Kang et al. (2025)\\\small\href{https://github.com/kangxin/NCAI}{Codebase} & \href{https://aclanthology.org/2025.neusymbridge-1.8.pdf}{Neuro-Conceptual Artificial Intelligence: Integrating OPM with Deep Learning to Enhance Question Answering Quality} & introduceNeuro-ConceptualArtificial Intelligence(NCAI),aspecializationof the neuro-symbolicAI approach that integrates conceptual modeling usingObject-Process Methodology (OPM) ISO19450:2024with deeplearningtoenhancequestion-answering (QA)quality \\
\hline
    \small\raggedright Yang et al. (2023)\\\small\href{https://github.com/azreasoners/recurrent_transformer}{Codebase} & \href{https://openreview.net/forum?id=udNhDCr2KQe}{Learning to Solve Constraint Satisfaction Problems with Recurrent Transformer} & Constraint satisfaction problems (CSPs) are about finding values of variables that satisfy the given constraints. We show that Transformer extended with recurrence is a viable approach to learning to solve CSPs in an end-to-end manner, having clear advantages over state-of-the-art methods such as Graph Neural Networks, SATNet, and some neuro-symbolic models. \\
\hline
    \small\raggedright Yang et al. (2023)\\\small\href{https://github.com/azreasoners/cl-ste}{Codebase} & \href{https://proceedings.mlr.press/v162/yang22h/yang22h.pdf}{Injecting Logical Constraints into Neural Networks via Straight-Through Estimators} & This paper introduces CL-STE, a method to inject discrete logical constraints into neural networks by representing the constraints as a loss function and using a Straight-Through Estimator (STE) to enable gradient-based optimization. By leveraging GPUs and avoiding heavy symbolic computation, this technique scales significantly better than existing neuro-symbolic methods and allows various architectures, like CNNs and GNNs, to learn from constraints with fewer or no labeled data. \\
\hline
    \small\raggedright Li et al. (2025)\\\small\href{https://github.com/Lizn-zn/NeqLIPS}{Codebase} & \href{https://arxiv.org/pdf/2502.13834}{Proving Olympiad Inequalities by Synergizing LLMs and Symbolic Reasoning} & we introduce a neuro-symbolic tactic generator that synergizes the mathematical intuition learned by LLMs with domain-specific insights encoded by symbolic methods. \\
\hline
\end{longtable}
\end{onecolumn}

\endgroup

\end{document}